\documentclass{bmvc2k}

\usepackage{graphicx}
\usepackage{amsmath}
\usepackage{amssymb}
\usepackage{booktabs}
\usepackage{adjustbox}
\usepackage{xcolor}
\usepackage{pgf} 
\usepackage{multirow} 
\usepackage{mathtools}
\usepackage{pifont}
\usepackage{tabularx}

\usepackage{algorithm}
\usepackage{algorithmic}
\newcommand{\RComment}[1]{\hfill{\color{green!60!black}$\triangleright$\,#1}}

\newcommand{\Up}{\textcolor{green!60!black}{$\uparrow$}}
\newcommand{\Down}{\textcolor{red!60!black}{$\downarrow$}}
\newcommand{\Same}{\textcolor{orange!50!black}{$\circ$}}
\newcommand{\red}[1]{\textcolor{red}{#1}}
\newcommand{\blue}[1]{\textcolor{blue}{#1}}

\newcommand{\teal}[1]{\textcolor{teal}{#1}}

\def\r{\red}
\def\b{\blue}

\def\s#1{}

\def\t{\teal}

\title{Zero-Shot Anomaly Detection with Dual-Branch Prompt Selection}

\addauthor{Zihan Wang}{zihan.wang5@mail.mcgill.ca}{1,2}
\addauthor{Samira Ebrahimi Kahou}{samira.ebrahimikahou@ucalgary.ca}{2,3,4}
\addauthor{Narges Armanfard
}{narges.armanfard@mcgill.ca}{1,2}

\addinstitution{
  McGill University \\
  Montreal, QC, Canada
}
\addinstitution{
  Mila – Quebec AI Institute \\
  Montreal, QC, Canada
}

\addinstitution{
  University of Calgary \\
  Calgary, AB, Canada
}
\addinstitution{
  CIFAR AI Chair \\
  Canada
}

\runninghead{WANG ET AL.}{DUAL-BRANCH PROMPT LEARNING FOR ZSAD}

\begin{document}

\maketitle

\begin{abstract}
    Zero‑shot anomaly detection (ZSAD) enables identifying and localizing defects in unseen categories by relying solely on generalizable features rather than requiring any labeled examples of anomalies. However, existing ZSAD methods, whether using fixed or learned prompts, struggle under domain shifts because their training data are derived from limited training domains and fail to generalize to new distributions. In this paper, we introduce PILOT, a framework designed to overcome these challenges through two key innovations: (1) a novel dual-branch prompt learning mechanism that dynamically integrates a pool of learnable prompts with structured semantic attributes, enabling the model to adaptively weight the most relevant anomaly cues for each input image; and (2) a label-free test-time adaptation strategy that updates the learnable prompt parameters using high-confidence pseudo-labels from unlabeled test data. Extensive experiments on 13 industrial and medical benchmarks demonstrate that PILOT achieves state-of-the-art performance in both anomaly detection and localization under domain shift.   
\end{abstract}
    
\section{Introduction}
\label{sec:intro}

Anomaly detection and localization is critical in safety-critical applications, including industrial inspection and medical diagnostics, where timely identification of rare deviations from normal patterns can prevent costly failures and improve patient outcomes. Traditional anomaly detection methods include supervised approaches that require labeled anomalous examples and unsupervised techniques relying solely on normal data~\cite{bovzivc2021mixed, hojjati2023dasvdd, li2021cutpaste, roth2022towards, karami2025graph, lai2023open}. Supervised methods require scarce and diverse anomaly labels, whereas unsupervised approaches avoid any need for anomalous samples by modeling only normal data, though they depend on comprehensive, representative normal data from the target domain~\cite{chandola2009anomaly}. Zero‑shot Anomaly Detection (ZSAD) has emerged as a powerful approach by leveraging pretrained vision–language models (VLMs) such as CLIP~\cite{clip}, which learns joint representations of images and text from large-scale image–text data, to identify anomalies in visual data without requiring any target-domain annotations of normal or anomalous samples.

\s{Prof Comment: detect anomalies target‑domain annotations of normal or anomalous samples  .... WHAT do you mean here?}
\s{Prof Comment: SUCCessful or prominent or attractive.... just to stay on the safe side of missing any prior non-LLM works }

Early successful ZSAD approaches utilized CLIP by designing explicit textual prompts that described both the object category and its condition, such as ``normal bottle” or ``anomalous transistor”. By crafting a large number of prompts that described relevant combinations of object category and condition, these methods enabled zero-shot classification and significantly outperformed vanilla CLIP baselines \cite{winclip, clipad, zoc}. These approaches are known as fixed prompt methods because they rely on manually designed textual prompts. More recent methods have shifted towards a single learnable prompt, where the prompt itself is represented as a trainable embedding and optimized during training. This optimization is performed using auxiliary data, which refers to anomaly detection datasets that are separate from the target domain. For example, AnomalyCLIP~\cite{AnomalyCLIP} learns single object agnostic prompt embeddings to capture generic anomaly patterns, while AdaCLIP~\cite{AdaCLIP} further integrates visual knowledge into the prompt embeddings to enhance model adaptation. 

\s{Prof Comment: Nice Job in this paragraph and the one before, except for the minor comments I gave.   Thank you!  }
\s{Prof Comment: SUCCessful or prominent or attractive.... just to stay on the safe side of missing any prior non-LLM works }

\s{Prof Comment: quickly explain what you mean here by domain shift--perhaps in the above paragraph} \s{Prof Comment: or maybe below? you almost touch on it below just have the word to clarify what do we mean by the domain shift-- just to clear the term since we also have target domain ...  }  

Despite these advances, both fixed and single learnable prompt approaches remain vulnerable to domain shift, which refers to performance degradation when data distribution in the target domain differs significantly from training data. Although pretrained VLMs like CLIP have strong generalization ability, their performance can drop sharply when there are significant differences between the data seen during pretraining and the target environment~\cite{bose2024stylip, feng2024rethinking}. Even when learnable prompt embeddings are further trained on auxiliary anomaly detection datasets, that are different from the actual deployment domain, these models often struggle to represent the full diversity of possible anomalies or adapt to new categories encountered at test time. This limited generalization occurs for two main reasons: first, the auxiliary and target datasets may have significant distribution differences, as we empirically demonstrate in Appendix~\ref{app:domain_gap}; second, single prompt embeddings risk overfitting to the patterns and features specific to the auxiliary dataset, which further reduces their ability to handle unseen anomalies. As a result, prompts derived from narrowly defined auxiliary sources may fail to capture diverse visual features in the target data~\cite{marzullo2024exploring}. Therefore, to address these challenges, we introduce a framework that learns and dynamically weights multiple prompts, rather than relying on a fixed or single learnable prompt.
\s{Prof Comment:  Therefore, \textit{we aim to develop a robust ZSAD framework that maintains robust performance under domain shift without access to any labeled target domain data.}} 
\s{Prof Comment: I think, THIS sentence can be moved down ... i will comment on the proper location once i reach there. So, perhaps you may instead  have a sentence here mentioning that one approach we take to address this problem is to allow the ZSAD model to learning multiple prompts?    then at the end of next paragraph, mention that to further manage the domain shift issue we leverage TTA ...} \s{Prof Comment: Good job here as well, except the minor comment -- thank you!}

\s{Though TTA aims to mitigate the domain shift issue between the train and test data, naive TTA strategies, \b{without customizing it for the specific task in hand,} \s{ in general vision tasks} can degrade performance when applied blindly, as the adaptation process may optimize proxy objectives that are not directly aligned with the true detection or localization goals, often resulting in suboptimal results~\cite{sun2020test, liu2021ttt, liang2025comprehensive}.}
On the other hand, Test-Time Adaptation (TTA) techniques have demonstrated improved model robustness in various computer vision tasks by refining model parameters on unlabeled test data~\cite{nado2020evaluating, liang2020we, schneider2020improving}. 
Although TTA is designed to address domain shift between training and test data, applying naive TTA strategies -- without tailoring them to the specific task at hand -- can inadvertently degrade performance. When used blindly, especially in tasks such as detection or localization, the adaptation process may optimize surrogate objectives that are misaligned with the true goals, often leading to suboptimal outcomes~\cite{sun2020test, liu2021ttt, liang2025comprehensive}. AnovL~\cite{deng2023anovl} was among the first to introduce TTA for anomaly localization refinement in ZSAD scenarios, but its focus is limited to localization and relies on fixed prompts. These limitations motivate us to explore how both anomaly detection and localization can be improved under domain shift by effectively utilizing unlabeled target data. \footnote{We present the Related Works section in Section A of the Supplementary Material.}.

\s{Prof Comment: I believe the above parts can be written much nicer, as the current presentation is not straightforward and does not direct us to the existing gap that we are trying to fill.  One paragraph on AD and ZSAD introduction in general. Then present the different existing categories in ZSAD, concluding with prompt-based models. Then present that these can be categorized into fixed/learnable... then what are the drawbacks of each... and presenting the research question you are trying to respond to here in this paper... then a paragraph on TTA, drawbacks of existing methods ending with your research question to be addressed by PILOT...  }

\s{Prof Comment: able  DO WE have any proof for showing that every prompt is actually specialized in a specific anomaly -- i believe we are just hoping for it to happen-- use a better word instead of able ...          however, Fig 2 cannot be considered as a proof/ it is just something that can be attributed to different anomalies. Do you have any figures showing how diverse are different prompts? e.g. a diversity map per dataset no, i don't have it, but maybe i can try to generate Yes it is needed to show that prompts are different and diverse that could be associated to different types of anomalies. can we visualize the prompts? for example showing their relevant texts by mapping them back to the language space i can try to compute similarity with the pretrained text embedding?   your prompts are embedding right? Yes  so, one experiment is to calculate the distance between final prompts within every dataset and also perhaps in the auxiliary dataset...       second experiment is to map back the embeddings to language using pretrained models just to show what each prompt has in human language      what do you think? The second experiment is worth to do, i am still thinking the first experiment, what's the point to calculate the distance for each prompt with final prompt within every dataset?--- you calculate the distances of every prompts vs all plot the histogram and i would like to see (still thinking how to visualize it the point is to show non-zero similarities between prompts     we need to find a proper way to show it in the embedding space      I guess what I would like to see is to see the distribution of distances. in total, we will have n(n-1)/2 numbers, and you need to plot the distribution of these values-- (combination of 2 out of n) n is number of prompts       then what i expect to see is a distribution not centering around zero        do you get it? this is the simplest way to show non-zero distance....  -- yes, i got it perfect  so then we can say here that as is demonstrated in our extensive experiments ..... got it?  Yes  let me know if you have any questions or still not clear.... It's clear) Let me think about it and firstly to try to see what happened} 
\s{Prof Comment: attribute memory bank     DEFINE what is memory bank and why it is relevant and useful, describe in lay language... perhaps you can refer to these terminologies in the previous paragraphs so that when you talk about them here all is clear}
To address these challenges, we propose a novel framework, \textbf{\underline{P}}rompt Learning with \textbf{\underline{I}}ntegrated 
\textbf{\underline{L}}abel‑Free Test‑Time Adaptation for Zer\textbf{\underline{O}}‑Sho\textbf{\underline{T}} Anomaly Detection (PILOT),  
a framework designed to mitigate the limitations of current ZSAD methods under domain shift. Instead of relying on a single learnable prompt, PILOT constructs a pool of learnable prompts, allowing the prompts to adaptively learn from various anomalies present in the auxiliary dataset. \s{Prof Comment: with each prompt specializing in different anomaly characteristics     WE DO NOT HAVE ANY PROOF HERE, as was discussed before -- Just a quick question here: as we provide the experiment, could i say this sentence? which one exactly? --  with each prompt specializing in different anomaly characteristics This one    I suggest to write it as allowing the prompts to adaptively learned to various anomalies existing in the data  something soft that we do not have explicit design on this  rather it is implicit   -- Okay, i will revise it to make it safe to say} This design enables the model to adaptively weight the most relevant prompts for each input image, allowing for more nuanced representations of diverse anomalies and supporting more effective TTA. During inference, our label-free test-time adaptation strategy updates all learnable prompts according to their alignment with each test image, using high-confidence pseudo-labels generated from unlabeled target data. This targeted adaptation facilitates rapid and reliable adjustment to new domains (i.e., the test domain) without requiring any labeled data. In addition, to mitigate overfitting to the auxiliary dataset and enhance adaptation stability, PILOT incorporates an attribute memory bank, which consists of a curated set of fixed prompts that describe normal and anomalous states, leveraging pretrained semantic knowledge from the text encoder. This attribute memory bank serves as a semantic anchor, helping to stabilize localization performance during TTA. Together, these innovations enable PILOT to achieve robust anomaly detection and localization across a variety of challenging real-world scenarios. Our contributions are summarized as follows:
\begin{itemize}
\item We propose a dual-branch prompt learning framework that dynamically weight
the most relevant prompts from both learnable prompt pool and attribute memory bank, enabling the model to leverage the most relevant anomaly cues for each input image.
\item To our knowledge, this is the first label‑free TTA strategy in ZSAD that leverages high‑confidence pseudo‑labels from unlabeled test samples to rapidly adapt prompt parameters at inference, yielding improvements in both anomaly detection and localization under domain shift.
\item Extensive evaluation on industrial and medical benchmarks shows that our method outperforms existing approaches in both anomaly detection and localization metrics.
\end{itemize}

\section{Proposed Method}
\label{sec:method}

ZSAD involves detecting and localizing anomalies in images from categories not encountered during the training phase. Formally, in the \emph{training phase}, we employ an auxiliary dataset defined as $\mathcal{D}_{\mathrm{aux}} = {(I_i, y_i, G_i)}_{i=1}^{|\mathcal{D}_{\mathrm{aux}}|}$, where each image $I_i\!\in\!\mathbb{R}^{H \times W \times 3}$ is paired with an image-level label
$y_i\!\in\!\{0,1\}$ (0~denotes normal and 1~denotes anomalous) and a pixel-level anomaly mask

$G_i\!\in\!\{0,1\}^{H \times W}$.
In the \emph{inference phase}, we consider an unlabeled target dataset $\mathcal{D}_{\mathrm{tar}} = \bigl\{I_t\bigr\}_{t=1}^{|\mathcal{D}_{\mathrm{tar}}|},$
where each $I_t \in \mathbb{R}^{H \times W \times 3}$. For every $I_t$, the model outputs an anomaly score $ \hat{y}_t \in [0,1],$ (with values closer to 1 indicating higher abnormality likelihood), along with a $\hat{G}_t \in [0,1]^{H \times W}$ that spatially localizes potential anomalous regions at pixel level. \s{Prof Comment: HERE mention that 0 means normal and 1 is anomaly so the closer to 1 means higher chance of being anomaly...}Crucially, the category sets of $\mathcal{D}_{\mathrm{aux}}$ and $\mathcal{D}_{\mathrm{tar}}$ are strictly disjoint, ensuring that the target domain remains unseen during training and thus establishing a true zero-shot setting. \s{Prof Comment: better to say ensuring for example zero-shot realization that the target domain has not been seen during the training phase.}

\s{Prof Comment: THIS SECTION might be unnecessary as the general idea is supposed to be presented at the end of the intro section. I felt a bit uncomfortable reading this again as I was expecting a step by step method in this section. That being said, we can have a summary of the method at the end of the Proposed method section including a flow chart... we may keep it for now but at least the first read was not pleasant.. }

\begin{figure*}[h]
    \centering
    \includegraphics[width=\linewidth]{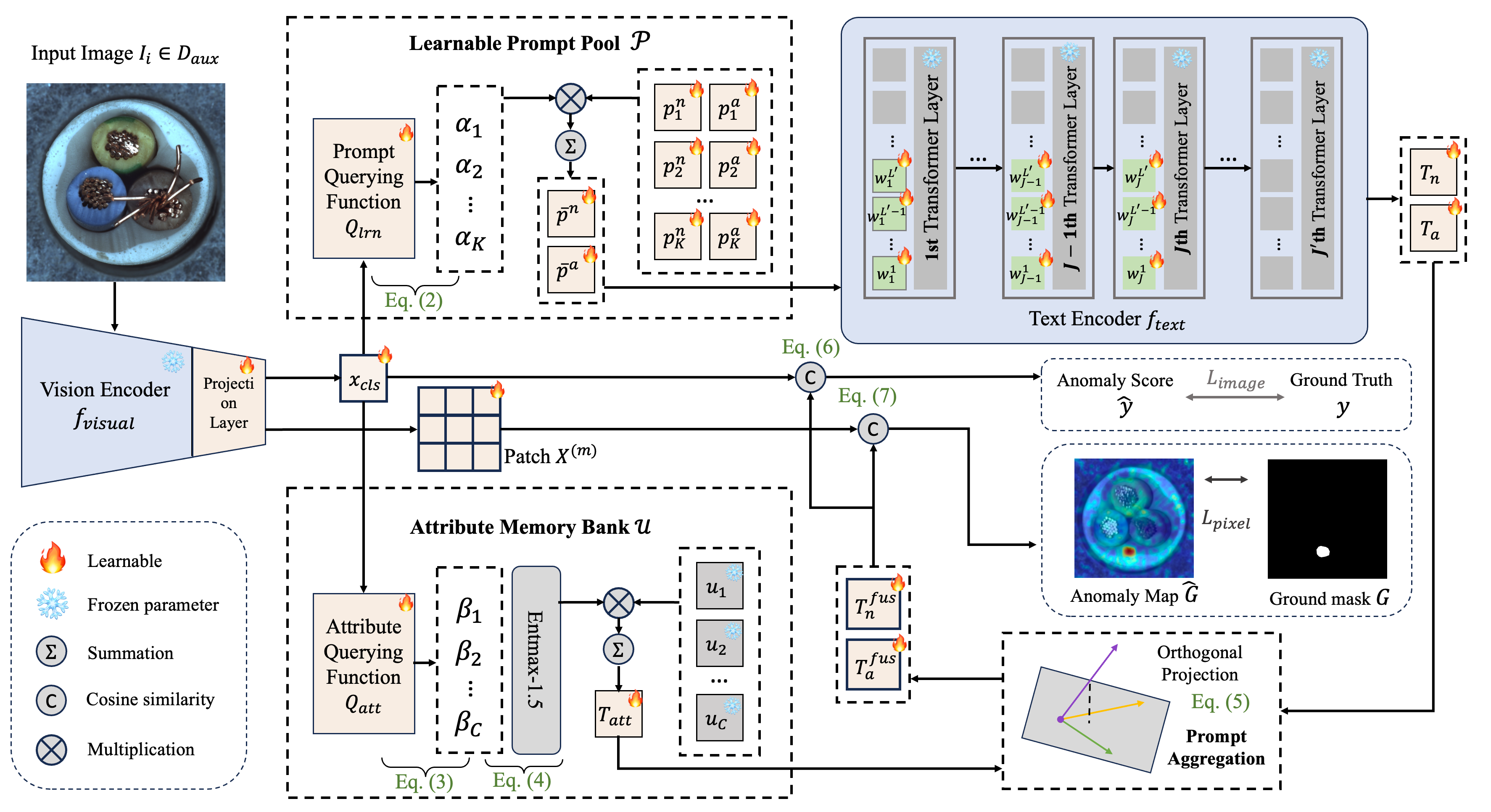}
    \caption{Overview of PILOT's training phase. The left panel shows the main workflow. The right panel details the Learnable Prompt Pool $\mathcal{P}$ and Attribute Memory Bank $\mathcal{U}$.}
    \label{fig:model}
\end{figure*}

\s{The block diagram of the proposed training phase of PILOT framework, highlighting the \b{Learnable Prompt Pool} $\mathcal{P}$ and Attribute Memory Bank $\mathcal{U}$, is shown in Fig.~\ref{fig:model}, and the inference phase is detailed in Section~\ref{method:TTA}. Details of each component are described in the following sections. For the
sake of completeness, we provide the pseudocode in Appendix~\ref{app:Pseudo_code}.}

To address the limitations of existing approaches, we propose a dual-branch framework that combines a learnable prompt pool $\mathcal{P}$ with an attribute memory bank $\mathcal{U}$. The training phase of our PILOT framework, depicted in Fig.~\ref{fig:model}, highlights these two key components. Further details on the inference phase are provided in Section~\ref{method:TTA}, with in-depth descriptions of each component in the following sections and the complete pseudocode in Appendix~E.

\subsection{Learnable Prompt Pool}
\label{method:learnable-prompt}

\s{Prof Comment: Throughout this section, I want you to mention that following.... we use e.g. a pair of learnable prompts pa and pn, however, to alleviate the [e.g. domain shift issue] we propose to include [e.g. sk and rk. These aim to ....]    the current format mixes the novel parts of PILOT with other existing things and  a reviewer could quickly reject the paper as they feel the paper is claiming existing things... being clear is always appreciated compared to being vague with many not actual claims -- i get it, thanks for your advise}

A successful recent approach for CLIP-based ZSAD employs a \emph{single} learnable prompt, defined as a pair of sequences $(p^n, p^a)$ representing normal and anomalous states, respectively, for binary anomaly classification~\cite{AnomalyCLIP}. Formally, this prompt can be represented as:
\begin{equation}
p^{n} = [v^{1},\! \dots,\! v^{L},\! \text{``[normal state] object''}],\;
p^{a} = [v^{1},\! \dots,\! v^{L},\! \text{``[anomalous state] object''}]
\end{equation}
where $L$ is the number of learnable prefix vectors, and each $v^{(i)} \in \mathbb{R}^d$ is a $d$-dimensional vector trained jointly with the model. These vectors form a “soft prompt,” intuitively acting as special tokens whose values steer the model's attention without modifying its core parameters, follows the prefix-tuning paradigm, which guides VLMs while keeping the main model weights frozen~\cite{li2021prefix, khattak2023maple}. The placeholders ``[normal state]” and ``[anomalous state]” filled with descriptors of the object’s condition, such as ``flawless'' and ``damaged.''

\s{Prof Comment: Yes, the brown is written well, except the minor comments i gave. goodjob! -- i will write following section like this! good. avoid redundant info that are already presented in the Intro, but always give clues and brief referring and reminders to help the reader continue reading without too much back and forth over the paper. -- thank you for your tips.}
\s{You should mention that within Qlrn, rk and sk are the learnable components-- i.e. specify that r and s are learnable. previously you had them in the figure and was easier to follow the role of s and r,  -- yes, i will revise it, to say it's learnable }
\s{, respectively, associated with $(p_k^n, p_k^a)$  WHAT do you mean here? respectively is needed? -- no, it's not needed, i just want to say each $(p_k^n, p_k^a)$ has corresponding $s_k$ and $r_k$ okay revise and here is the place to carefully define rk and sk -- yes, i will revise it} \s{make sure this blue part is correct and in line with what you will write -- sorry.. you mean brown part? yes}
Despite its simplicity and effectiveness, relying solely on a single learnable prompt for all input images limits generalization under domain shift, as discussed in the introduction, because a single pair of prompts often fails to adequately represent the diversity of anomaly patterns encountered in different scenarios. To address this limitation, we propose extending this single prompt to a learnable prompt \textit{pool}, containing multiple learnable prompts, each associated with additional embeddings for adaptive weighting and alignment. Formally, the proposed prompt pool is defined as \(\mathcal{P} = \{(p_{k}^{n},\,p_{k}^{a},\,s_k,\,r_k)\}_{k=1}^{K}\), where \(K = |\mathcal{P}|\) is the cardinality of the prompt pool. Each $(p_k^n, p_k^a)$ is constructed similarly to previous works, and the additional learnable embeddings $s_k$ and $r_k$ enable adaptive weighting of the prompts and alignment with the input image features. Here, $s_k$ is a modulation vector for prompt $k$, applied element-wise to the image feature $x_{\mathrm{cls}}$ to adaptively emphasize information relevant to that prompt, while $r_k$ provides a prompt-specific reference for measuring similarity. These embeddings facilitate an adaptive emphasis on prompts that best align with the diverse patterns present in the auxiliary dataset. To achieve this adaptive prompt weighting, we introduce a prompt-querying function, which computes prompt relevance scores given an image feature $x_{\mathrm{cls}} \in \mathbb{R}^d$ produced by the CLIP vision encoder, i.e. $f_{\mathrm{visual}}$, from an image $I_i$ in 
$\mathcal{D}_{\mathrm{aux}}$,:
\begin{equation}
Q_{\mathrm{lrn}}\bigl(x_{\mathrm{cls}},\, \mathcal{P}\bigr) = \bigl\{ \alpha_k \coloneqq \mathrm{sim}\bigl(x_{\mathrm{cls}} \odot s_k,\, r_k\bigr) \bigr\}_{k=1}^{K},
\label{eq:prompt-query-main}
\end{equation}
where “$\odot$” denotes element-wise multiplication, and $\mathrm{sim}(\cdot,\cdot)$ represents cosine similarity. Intuitively, \(\alpha_{k}\) quantifies how well the image content aligns with the k-th prompt pair; this is realized by weighting the image feature $x_{\mathrm{cls}}$ with the attention mask $s_k$ and measuring its similarity to the associated reference embedding $r_k$. The higher the score \(\alpha_{k}\) is, the more similar the input image is to the k-th prompt $(p_{k}^{n},\,p_{k}^{a})$. These scores are then used to form weighted combinations of the prompt token sequences as
\(\bar p^n = \sum_{k=1}^K \alpha_k p_k^n\), \(\bar p^a = \sum_{k=1}^K \alpha_k p_k^a\).
The resulting \(\bar p^n\) and \(\bar p^a\) are subsequently encoded using the CLIP pretrained text encoder to obtain the final text embeddings
\(T_n = f_{\text{text}}(\bar p^n)\) and \(T_a = f_{\text{text}}(\bar p^a)\),
which are employed for downstream anomaly scoring and localization. This process is visualized in Figure~\ref{fig:model}.  

Although the text encoder $f_{\mathrm{text}}$ is frozen to preserve CLIP's rich pretrained textual representations, we employ a prefix-tuning strategy to enable effective fine-tuning and adaptation of CLIP. Inspired by similar prefix-tuning methods~\cite{khattak2023maple}, we introduce additional learnable parameters to $f_{\mathrm{text}}$. More specifically, as is shown in Figure~\ref{fig:model}, layer-wise prefix vectors are defined as
\(
\bigl\{\,w_j^{(\ell)} \in \mathbb{R}^d \mid j = 1,\dots,J,\; \ell = 1,\dots,L'\bigr\}
\), 
where $f_{\mathrm{text}}$ consists of a total of $J'$ transformer layers, and $J \leq J'$ denotes the number of initial layers to which learnable vectors are appended. $L'$ represents the number of learnable vectors per layer. Specifically, these prefix vectors are prepended as additional tokens to the input sequence at the first $J$ transformer layers, leaving the original transformer layers unchanged. At each layer $j$, these vectors \(\{w_j^{(\ell)}\}_{\ell=1}^{L'}\) (illustrated as green blocks within the $f_{\mathrm{text}}$ module of Figure~\ref{fig:model}) are added before the standard, frozen token embeddings (represented as gray blocks in the the $f_{\mathrm{text}}$ module of Figure~\ref{fig:model}) prior to the self-attention mechanism in in each frozen transformer layer. This approach enables adaptation of the model’s internal representations without modifying the pretrained CLIP parameters, thereby preserving their robustness and generalization capabilities. Furthermore, as is illustrated in Figure \ref{fig:model}, PILOT fine-tunes only the final $d$-dimensional projection layer within the vision encoder $f_{\mathrm{visual}}$. By restricting updates exclusively to this projection layer, we substantially reduce memory overhead and computational complexity. This targeted fine-tuning strategy also mitigates the risk of overfitting to auxiliary datasets, maintaining the strong generalization properties inherent to CLIP~\cite{fahes2024fine}.

\subsection{Attribute Memory Bank}
\label{method:attr-prompt}

\s{Prof Comment: Basically, it is using the Clip text encoder? Note that in Fig. 1 and other places you have mentioned that ftext includes the learnable parameters, different from the original text of CLIP. DO you get the point here? one could think that you complete training of your proposed text (including the new parameters) then use that for creating these attributes..... clarify  -- it's using original Clip text encoder, i think i want to mentioned in previous section that ftext is frozen, the extra prefix vectors (w) just append into ftext     But in the figure you are referring to the whole block, including learnable parameters, as ftext.... you may have ftextclip and ftextpilot?  ... yes i could, but i don't want o draw ftextclip into the diagram... two text encoder would be too much, maybe just mentioned ftextclip in here which is original clip text encoder? and draw ftextpilot in diagram include learnable parameters    I suggest to keep as is but clarify that ftext clip is the ones frozen (snow flake) shown in gray and we simply appended a few learnable blocks shown in orange (on fire) -- yes, i will revise it} 

Although the learnable prompt pool $\mathcal{P}$ aims to provide adaptive and diverse representations tailored to the auxiliary dataset, it may still be susceptible to overfitting when deployed on target domains. To mitigate this issue, we further introduce a fixed attribute memory bank comprising attribute embeddings extracted from the pretrained CLIP text encoder $f_{\text{text}}$ (prior to applying our modifications). Recent works, such as WinCLIP~\cite{winclip}, have demonstrated the robustness of fixed textual prompts by encoding multiple natural-language templates. Inspired by this, we construct our attribute memory bank from a set of descriptive natural-language templates (e.g., ``A bright photo of a [State] object”), where \textit{[State]} indicates descriptive attributes of the object’s condition (e.g., ``normal", ``corroded", or ``contaminated").

\s{Prof Comment: add the c=1 to C}
\s{Prof Comment:in the above equation, for wc, you need to say, c=1...C. }
Each attribute template is passed through the frozen CLIP text encoder $f_{\mathrm{text}}$, without any additional learnable tokens $w_j^{(\ell)}$, to produce a fixed attribute embedding, formally defined as \(\mathcal{U} = \bigl\{(u_c,\, g_c)\bigr\}_{c=1}^C,\) where $C$ is the number of attribute embeddings, and each $u_c \in \mathbb{R}^d$ represents the EOS-token embedding of the corresponding template. Unlike WinCLIP, which averages embeddings across multiple prompts, we associate each attribute embedding $u_c$ with a learnable modulation vector $g_c$,  similar to the learnable prompt pool. Similar to how $s_k$ adapts the image features for each prompt, $g_c$ dynamically modulates the contribution of each attribute according to the input image. Specifically, given an image feature $x_{\mathrm{cls}}$, we introduce the attribute-querying function to compute attribute relevance scores as below:
\begin{equation}
Q_{\mathrm{att}}\bigl(x_{\mathrm{cls}},\, \mathcal{U}\bigr) = \bigl\{ \beta_c \coloneqq \mathrm{sim}\bigl(x_{\mathrm{cls}} \odot g_c,\, u_c\bigr) \bigr\}_{c=1}^C,
\label{eq:attribute-query-merged}
\end{equation}
where “$\odot$” denotes element-wise multiplication, and $\mathrm{sim}(\cdot,\cdot)$ is the cosine similarity. Here, $u_c$ itself serves as the reference embedding for each attribute, as it is obtained directly from the pretrained $f_{\mathrm{text}}$ and already contains rich semantic information. Each score $\beta_c$ quantifies the alignment between the image and the $c$-th attribute embedding. The resulting scores $\{\beta_c\}_{c=1}^C$ quantify the alignment between the input and each attribute. Following~\cite{gumbel}, we inject i.i.d. Gumbel noise into the attribute logits and apply the $\operatorname{entmax}{1.5}$ operator~\cite{entmax} to obtain a sparse yet differentiable weighting over the most prominent cues:
\begin{equation}\label{eq:entmax}
    \{\phi_c\}_{c=1}^C = \operatorname{entmax}{1.5}(\beta_c + \text{noise}),
    \quad
    \sum_{c=1}^C \phi_c = 1,\quad
    \phi_c \ge 0.
\end{equation}
We then compute the attribute text embedding as a weighted combination.
\(
  T_{att} = \sum_{c=1}^C \phi_c u_c,{}\\ T_{att} \in \mathbb{R}^d
\) The injected noise promotes exploration and prevents premature convergence to a single attribute, allowing the model to capture diverse and complementary semantic cues. By providing stable, pretrained semantic cues, the attribute memory bank acts as a semantic anchor during TTA discussed in Section \ref{method:TTA}. This mechanism helps to counteracts the potential overfitting and instability of learnable prompts, ensuring the model retains reliable semantic references even as it adapts to new domains.
\subsection{Prompt Aggregation via Orthogonal Projection}
\label{method:aggregation}
\s{Prof Comment: where did you get this idea? just cite it -- yes, i will}
\s{Prof Comment: DO you have visualization of $\bigl( T_{(\cdot)} - \mathrm{Proj}_{T_{\mathrm{att}}}(T_{(\cdot)}) \bigr)$? I wonder how big/small are these values? This shows if Tn/a are actually provide any additional info to Tatt. WHat I would like to see is histogram of these and hopefully they are not all zero! DO you have any idea of these values? have you tracked them in your implementation? -- no, but it's obvioulsy to see they are different, as Tn/Ta are finetuned on auxiliary dataset, and Tatt are from pretrained clip    Well, i see it as an ablation in support of the need to having this learnable prompt beside the pretrained clip outputs note that you have a selection mechanism applying on the clip output as well that is another level of fine tuning. -- that's can be another ablation study i agree   Please include as I (as a reviewer) am not yet fully convinced that the two prompts are really needed.... this ablation helps to show that we are adding info to the system by including P. -- Okay, let me add another experiment}
\s{Wouldn't be better to use something different than $\phi$ here as you already have it above in (7), -- you are right, i got it, i will revise the notation for projection formula yes}
Following~\cite{yang2021dolg}, to robustly leverage both the stable semantics from the attribute embedding $T_{att}$ and the adaptive semantics from the learnable prompts ($T_n$ for normal and $T_a$ for anomalous), we aggregate these representations using an orthogonal projection operation instead of simple averaging or concatenation. For each prompt embedding $T_{(\cdot)}$ (where $(\cdot) \in \{n, a\}$), we decompose it into components parallel and orthogonal to $T_{att}$:
\begin{equation}
T_{(\cdot)}^{\mathrm{fus}} = T_{\mathrm{att}} + ( T_{(\cdot)} - \mathrm{Proj}_{T_{\mathrm{att}}}(T_{(\cdot)}) ),\quad
\mathrm{Proj}_{\eta}(\psi) = \frac{\eta^\top \psi}{\|\eta\|^2}\,\eta
\label{eq:fuse-main}
\end{equation}
This construction anchors the fused embedding in the robust attribute space, while enriching it with only those cues from the learnable prompt that are \emph{complementary} to the attribute. Geometrically, $T_{(\cdot)}^{\mathrm{fus}}$ preserves all the core semantics provided by $T_{att}$, and selectively incorporates novel information from $T_{(\cdot)}$ that lies in the orthogonal complement. Further analysis can be found in Appendix~\ref{app:orthogonal_projection_tracker}.
This parameter-free aggregation prevents redundancy and enhances robustness to distributional shift, especially during TTA, as shown in Table~\ref{table:Projector_ablation}.

\s{Prof Comment:enhances robustness to distributional shift        WHY Such a claim?  any support? -- yes, i have ablation study, that using with parameter method degrade pixel level performance   WHICH section? Prompt Aggregation Ablation. section 4.4    I don't see any support to this claim there! how does such aggregation enhance robustness to domain shift? -- i tried other two Prompt Aggregation method with parameter but didn't show better results, the cross-attention has a lot of more paramters, has better image level performance but degrade the pixel level  Which figure/table are you talking about? Table 6 I missed it  let me see, Table~\ref{table:Projector_ablation}  Good thanks for clarifying -- you are welcome, i will make the table with clear definition    I suggest to directly refer to the table  here to make things clear rather than only section 4.4.   -- Okay, i will revise it }

\subsection{Training Objectives}
\label{method:training_obj}

To achieve both accurate anomaly detection and precise localization, we jointly optimize image-level classification and pixel-level segmentation objectives. Joint training allows the two tasks to reinforce each other, as global context aids localization while fine-grained cues improve classification. All training is performed on the auxiliary dataset $\mathcal{D}_{aux}$. Given the aggregated text embedding ${T}_{(\cdot)}^{fus}$ and the global visual feature $x_{\mathrm{cls}}$, we compute the anomaly score and corresponding image-level loss:
\begin{equation}
\hat{y} = \sigma\bigl(\mathrm{sim}(x_{\mathrm{cls}},\, T_{(\cdot)}^{\mathrm{fus}})\bigr),\quad
\mathcal{L}_{\mathrm{image}} = \mathrm{BCE}(\hat{y},\, y)
\label{eq:image-score}
\end{equation}
where $\sigma(\cdot)$ denotes the sigmoid function, $\mathrm{sim}(\cdot,\cdot)$ is cosine similarity, and $\mathrm{BCE}$ is the binary cross-entropy loss comparing the predicted score $\hat{y}$ to the ground-truth image label $y \in \{0,1\}$. For pixel-level localization, following~\cite{AnomalyCLIP}, we extract patch features
\(\{\,X^{(m)}\in\mathbb{R}^{H\times W\times d}\}_{m\in\mathcal{M}}\)
from a set $\mathcal{M}$ of selected layers in $f_{\mathrm{visual}}$. Each $X^{(m)}$ is formed by reshaping the patch token sequence into a spatial grid of size $H \times W$ with $d$-dimensional embeddings. For each $m\in\mathcal{M}$, we compute an anomaly map using the aggregated text embedding:
\begin{equation}
\hat{G}^{(m)} = \sigma\bigl(\mathrm{sim}(X^{(m)},\, T_{(\cdot)}^{\mathrm{fus}})\bigr),\quad
\mathcal{L}_{\mathrm{pixel}} = \frac{1}{|\mathcal{M}|}
\sum_{m\in\mathcal{M}}
\bigl[
  \mathrm{Focal}(\hat G^{(m)},\, G) +
  \mathrm{Dice}(\hat G^{(m)},\, G)
\bigr]
\label{eq:pixel-loss}
\end{equation}
where $\hat G^{(m)} \in [0,1]^{H \times W}$ is the predicted anomaly probability map for layer $s$, $G$ is the ground-truth pixel-level mask, $\mathrm{Focal}(\cdot,\cdot)$ denotes the focal loss~\cite{focal}, and $\mathrm{Dice}(\cdot,\cdot)$ denotes the Dice loss~\cite{dice}. This combination balances pixel-wise discrimination and overlap accuracy. The final training objective combines both terms \(\mathcal{L}_{\mathrm{PILOT}} = \mathcal{L}_{\mathrm{image}} + \mathcal{L}_{\mathrm{pixel}}.\)

\subsection{Test-Time Adaptation}
\label{method:TTA}
A central challenge for robust ZSAD, as outlined in the Introduction, is achieving reliable adaptation to target domains that may differ significantly from the auxiliary dataset. While fine-tuning on $\mathcal{D}_{\mathrm{aux}}$ can calibrate model representations, substantial domain shifts often cause conventional approaches to overfit or fail to generalize, especially in the absence of labeled data from the target domain. To overcome these obstacles, we propose a TTA strategy that enables the model to adjust to new domains using only unlabeled target data, without compromising the integrity of previously learned representations.

\s{Prof Comment: The core idea of our TTA scheme is to leverage high-confidence pseudo-labels as surrogates for ground-truth annotations in the target domain. IS THIS ACCURATE?  we don't use high confidence labels for annotation, we rather use them for tweaking of parameters -- use high-confidence pseudo-labels to updated the parameters, Yes, in TTA, the pseudo label replace the actual label, so for me it can be consider as surrogates for ground-truth annotations?  WRONG! you do not re-lable all test samples, you only label reliable samples and use them for model updates, right?  -- Yes-- so you do not have surrogate for all samples! basically you have no surrogate, you just identify the high-confidence samples, use them for model update-- number of samples used for update depend on $\rho$, right? -- right, maybe i misunderstand the word surrogates, i will revise it, thanks for pointing out!  } 
The core idea of our TTA strategy is to identify a subset of highly reliable predictions (high-confidence pseudo-labels) from the target dataset and leverage these predictions to update model parameters. Specifically, for each image $I_t$ in the unlabeled target dataset $\mathcal{D}{\mathrm{tar}}$, we compute the anomaly score as $\tilde{y}t = \sigma\big(\mathrm{sim}(x{\mathrm{cls}}^t,, T{(\cdot)}^{\mathrm{fus}})\big)$, where $x_{\mathrm{cls}}^t$ is the CLIP visual feature for $I_t$ and $T_{(\cdot)}^{\mathrm{fus}}$ denotes the fused prompt embedding selected and aggregated for $I_t$ using model parameters trained on $\mathcal{D}_{\mathrm{aux}}$. We then identify the most reliable samples by selecting the top and bottom $\rho$-fractions of scores:
\begin{equation}
A = \mathrm{Top}_{\lceil \rho B \rceil} \{\tilde{y}_t\}_{t=1}^B, \qquad
N = \mathrm{Bot}_{\lceil \rho B \rceil} \{\tilde{y}_t\}_{t=1}^B,
\end{equation}
where $B = |\mathcal{D}_{\mathrm{tar}}|$ and $0 < \rho < 0.5$. Pseudo‐labels are then defined by \(\hat y_t = 1\) if \(t \in A\), and \(\hat y_t = 0\) if \(t \in N\). Adaptation is performed by optimizing an image-level binary cross-entropy loss on the selected samples, i.e., $\mathcal{L}_{\mathrm{image}} = \mathrm{BCE}(\hat{y}_i, \tilde{y}_i)$ for $i \in A \cup N$.

\s{Prof Comment: In the above formula, write that $i\in (A U N)$--  yes, i will}

\s{Prof Comment: selected ?? do you have a threshold to select? the s is not binary if i recall correctly. -- i should not use select...}
Critically, during TTA, we update only the parameters associated with the learnable prompt pool, including the learnable prompts, their corresponding embeddings $s_k$ and $r_k$. All other model parameters, such as those in the attribute memory bank and the core CLIP architecture, remain fixed. The updates for each learnable prompt are weighted by the attention scores defined previously (see Section~\ref{method:learnable-prompt}), meaning prompts with higher attention scores for a given input contribute more significantly to parameter updates. We limit adaptation strictly to the image-level objective because, as shown in our experiments (Section~\ref{exp:TTA_effect}), pseudo-labels derived at the image level exhibit considerably higher reliability compared to those at the pixel level. Pixel-level pseudo-labeling can be adversely affected by spatial ambiguity and noise exacerbated by domain shifts. By focusing adaptation solely on image-level labels, we reduce the risk of propagating noisy pseudo-labels, ensuring stable, unsupervised adaptation in the target domain. Our experimental results demonstrate that applying such a strategy improves both pixel-level and image-level detection on the target domain.

\section{Experiments} 
\label{sec:experiments}

\noindent\textbf{Benchmarks.}
\s{Prof Comment: Note that the localization‑only medical sets contain only anomalous samples; hence, we report their results without TTA .... NOT clear what do you mean here. clarify. -- medical dataset only has anomaly data, no normal data, so my TTA cannot be applied  okay-- just clarify here. -- okay!}
To demonstrate the effectiveness of our method, we conduct experiments on 13 widely used benchmarks spanning industrial inspection and medical imaging. Specifically, we evaluate on six industrial datasets: BTAD \cite{dataset_btad}, VisA \cite{dataset_visa}, MVTec AD \cite{dataset_mvtec}, DAGM \cite{dataset_DAGM}, SDD \cite{dataset_ksdd}, and DTD‑Synthetic \cite{dataset_DTD} and seven medical datasets, subdivided into detection-only (HeadCT \cite{dataset_headCT}, BrainMRI \cite{dataset_brain}, Br35H \cite{dataset_br35h}) and localization (ISIC \cite{dataset_isic}, CVC ColonDB \cite{dataset_ColonDB}, CVC ClinicDB \cite{dataset_ClinicDB}, Kvasir \cite{dataset_Kvasir}). The medical datasets for anomaly localization contain only anomalous samples and lack normal data; since our TTA strategy requires normal samples, results are reported without TTA (see Appendix~H). To ensure strict zero-shot evaluation, auxiliary datasets exclude all target data. More specifically, for a fair comparison, we followed the related literature \cite{AnomalyCLIP, AdaCLIP}, and used MVTec AD~\cite{dataset_mvtec} as the auxiliary dataset when reporting the performance on all other datasets, while when reporting the performance on MVTec AD, we trained the model on VisA~\cite{dataset_visa} as the auxiliary dataset.

\noindent\textbf{Baseline.} We compare PILOT with the most recent SOTA methods, including SAA\cite{saa}, CoOp \cite{CoCoOp}, WinCLIP \cite{winclip}, APRIL-GAN \cite{April_GAN},
AnomalyCLIP \cite{AnomalyCLIP}, and AdaCLIP \cite{AdaCLIP}.

\noindent\textbf{Evaluation Metrics.} Following standard practices in zero-shot anomaly detection literature such as ~\cite{April_GAN, AnomalyCLIP}, we evaluated our method using the AUROC. Additionally, we report APs for anomaly detection and use AUPRO~\cite{aupro} to assess pixel-level anomaly localization performance. Results are provided at both dataset-level and domain-level averages.

\begin{table}[h]
    \centering
    \caption{Model performance on public benmarks. The top section reports image‑level results (AUROC, AP), and the bottom section reports pixel‑level results (AUROC, AUPRO). Best entries are marked with bold and second‑best entries with underline.}
    \label{table:industrial_results}
    \begin{adjustbox}{width=0.9\textwidth}
    \begin{tabular}{@{}lccccccc@{}}
        \toprule
        \multicolumn{8}{c}{\textbf{Industrial Image-level Anomaly Detection (AUROC, AP)}} \\
        \midrule
        Dataset        & SAA              & CoCoOp            & WinCLIP                   & APRIL-GAN         & AnomalyCLIP                    & AdaCLIP              & PILOT                  \\
        \midrule
        MVTec AD       & (64.2,  87.0)    & (88.2,  94.5)    & (\underline{91.9}, \underline{96.4}) & (86.5, 93.2)    & (91.6, 96.3)    & (91.3, 95.8)    & (\textbf{92.1}, \textbf{96.6})    \\[0.5ex]
        VisA           & (67.5,  76.2)    & (63.2,  68.5)    & (78.5,  81.4)             & (78.2, 81.3)    & (\underline{81.8}, \underline{85.1}) & (81.6, 84.1)    & (\textbf{84.1}, \textbf{85.7})    \\[0.5ex]
        BTAD           & (59.4,  89.2)    & (67.0,  77.6)    & (68.4,  71.1)             & (73.8, 68.9)    & (83.8, 88.4)    & (\underline{90.4}, \underline{90.7}) & (\textbf{95.3}, \textbf{96.8})    \\[0.5ex]
        SDD            & (68.9,  37.9)    & (75.1,  65.3)    & (84.5,  77.6)             & (80.0, 71.7)    & (\underline{85.5}, \underline{81.7}) & (81.4, 72.8)    & (\textbf{87.9}, \textbf{85.7})    \\[0.5ex]
        DAGM           & (87.3,  88.6)    & (87.7,  74.8)    & (91.9,  79.7)             & (94.6, 84.0)    & (96.1, 90.2)    & (\underline{97.3}, \textbf{93.1}) & (\textbf{98.5}, \underline{92.8})    \\[0.5ex]
        DTD-Synthetic  & (94.6,  93.7)    & (92.4,  94.8)    & (93.4,  92.8)             & (86.7, 95.2)    & (93.7, 97.3)    & (\underline{97.5}, \underline{98.2}) & (\textbf{97.7}, \textbf{98.4})    \\[0.5ex]
        \midrule
        Mean           & (73.7,  78.8)    & (78.9,  79.3)    & (84.8,  83.2)             & (83.3, 82.4)    & (88.8, \underline{89.8}) & (\underline{89.9}, 89.1)    & (\textbf{92.6}, \textbf{92.7})    \\
        \midrule
        \multicolumn{8}{c}{\textbf{Medical Image-level Anomaly Detection (AUROC, AP)}} \\
        \midrule
        HeadCT         & (46.8, 68.0)    & (78.4, 78.8)    & (81.8, 80.2)             & (89.1, 89.4)    & (\underline{91.0}, \underline{92.3}) & (86.5, 81.3)    & (\textbf{94.5}, \textbf{96.0})    \\
        BrainMRI       & (34.4, 76.7)    & (61.3, 44.9)    & (86.6, 91.5)             & (89.3, 90.9)    & (\underline{93.2}, \underline{94.7}) & (93.2, 89.7)    & (\textbf{97.3}, \textbf{97.9})    \\
        Br35H          & (33.2, 67.3)    & (86.0, 87.5)    & (80.5, 82.2)             & (93.1, 92.9)    & (\underline{95.3}, \underline{96.0}) & (92.1, 90.9)    & (\textbf{96.7}, \textbf{97.7})    \\
        \midrule
        Mean           & (38.1, 70.7)    & (75.2, 70.4)    & (83.0, 84.6)    & (90.5, 91.1)    & (\underline{93.2}, \underline{94.3}) & (90.6, 87.3)    & (\textbf{96.2}, \textbf{97.2})      \\
        \midrule
        \multicolumn{8}{c}{\textbf{Industrial Pixel-level Anomaly Localization (AUROC, AUPRO)}} \\
        \midrule
        MVTec AD       & (75.5, 38.1)   & (44.4, 11.1)   & (85.1, 64.6)   & (87.6, 44.0)   & (\textbf{90.7}, \underline{78.7})   & (89.4, 37.8)   & (\underline{90.3}, \textbf{80.2})   \\[0.5ex]
        VisA           & (76.5, 31.6)   & (42.1, 12.2)   & (79.6, 56.8)   & (94.2, \underline{86.8})   & (95.3, 85.1)   & (\underline{95.5}, 77.8)   & (\textbf{96.0}, \textbf{87.2})   \\[0.5ex]
        BTAD           & (65.8, 14.8)   & (28.1, 6.5)    & (72.7, 27.3)   & (60.8, 25.0)   & (94.4, \underline{73.6})   & (\underline{94.8}, 32.5)    & (\textbf{96.3}, \textbf{74.9})   \\[0.5ex]
        SDD            & (78.8, 6.6)    & (24.4, 8.3)    & (68.8, 24.2)   & (79.8, 65.1)   & (\underline{90.7}, \underline{66.6})   & (71.7, 17.6)   & (\textbf{92.4}, \textbf{73.4})   \\[0.5ex]
        DAGM           & (62.7, 32.6)   & (17.5, 2.1)    & (87.6, 65.7)   & (82.4, 66.2)   & (\underline{95.6}, \underline{91.0})   & (91.1, 62.3)   & (\textbf{96.0}, \textbf{91.4})   \\[0.5ex]
        DTD-Synthetic  & (76.7, 60.6)   & (14.8, 3.0)    & (83.9, 57.8)   & (95.3, 86.9)   & (97.8, \textbf{91.1})      & (\textbf{98.2}, 86.2)   & (\underline{98.0}, \underline{90.2})   \\[0.5ex]
        \midrule
        Mean           & (72.7,  30.7)    & (28.6,   7.2)    & (79.6,  49.4)             & (83.4, 62.3)    & (\underline{94.1}, \underline{81.0}) & (90.1, 52.4)    & (\textbf{94.8}, \textbf{82.9})    \\
        \bottomrule
    \end{tabular}
    \end{adjustbox}
\end{table}

\s{Prof Comment: Here in this table, you are using P as for pixel-level accuracy, while previously it was used for the prompt pool -- oh.. i didn't realize it, i will change a notation.. sorry about that...}
\begin{table}[h] 
  \centering
  \caption{Comparison of PILOT vs.\ state‑of‑the‑art on industrial benchmarks. $\mathbf{Img}$ and $\mathbf{Pix}$ denote image‑level and pixel‑level metrics, respectively. ``w/o TTA'' and ``w/ TTA'' indicate performance without and with TTA, respectively. Arrows indicate change in performance: \Up (increase $>0.5$), \Down (decrease $<-0.5$), and \Same (no significant change, $\pm0.5$).}
  \label{table:tta_ab}
  \begin{adjustbox}{width=\textwidth,center}
    \begin{tabular}{ll|cc|cc|cc|cc}
      \toprule
      \textbf{Dataset} &  
        & \multicolumn{2}{c|}{\textbf{APRIL-GAN}} 
        & \multicolumn{2}{c|}{\textbf{AnomalyCLIP}} 
        & \multicolumn{2}{c|}{\textbf{AdaCLIP}} 
        & \multicolumn{2}{c}{\textbf{PILOT}} \\
      \cmidrule(lr){3-4} \cmidrule(lr){5-6} \cmidrule(lr){7-8} \cmidrule(lr){9-10}
        & 
        & \textbf{w/o TTA} & \textbf{w/ TTA}
        & \textbf{w/o TTA} & \textbf{w/ TTA}
        & \textbf{w/o TTA} & \textbf{w/ TTA}
        & \textbf{w/o TTA} & \textbf{w/ TTA} \\
      \midrule
      \multirow{2}{*}{MVTec AD}      
        & $\mathbf{Img}$ & (86.5, 93.2) & (82.9\Down, 89.8\Down)
               & (91.6, 96.3) & (89.2\Down, 95.0\Down)
               & (91.3, 95.8) & (90.1\Down, 95.8\Same)
               & (90.2, 95.7) & (92.1\Up, 96.6\Up) \\
        & $\mathbf{Pix}$ & (87.6, 44.0) & (86.0\Down, 42.0\Down)
                & (90.7, 78.7) & (88.1\Down, 73.5\Down)
                & (89.4, 37.8) & (87.0\Down, 29.5\Down)
                & (89.4, 79.2) & (90.3\Up, 80.2\Up) \\
      \midrule
      \multirow{2}{*}{VisA}          
        & $\mathbf{Img}$ & (78.2, 81.3) & (84.5\Up, 88.0\Up)
               & (81.8, 85.1) & (84.3\Up, 85.8\Up)
               & (81.6, 84.1) & (83.0\Up, 85.0\Up)
               & (82.0, 84.1) & (84.1\Up, 85.7\Up) \\
        & $\mathbf{Pix}$ & (94.2, 86.8) & (92.0\Down, 84.0\Down)
                & (95.3, 85.1) & (85.5\Down, 82.6\Down)
                & (95.5, 77.8) & (93.8\Down, 72.6\Down)
                & (94.8, 85.6) & (96.0\Up, 87.2\Up) \\
      \midrule
      \multirow{2}{*}{BTAD}          
        & $\mathbf{Img}$ & (73.8, 68.9) & (76.9\Up, 73.2\Up)
               & (83.8, 88.4) & (85.7\Up, 93.0\Up)
               & (90.4, 90.7) & (91.6\Up, 92.5\Up)
               & (90.0, 91.6) & (95.3\Up, 96.8\Up) \\
        & $\mathbf{Pix}$ & (60.8, 25.0) & (58.0\Down, 23.0\Down)
                & (94.4, 73.6) & (94.1\Same, 72.0\Down)
                & (94.8, 32.5) & (93.0\Down, 25.5\Down)
                & (94.9, 73.0) & (96.3\Up, 74.9\Up) \\
      \midrule
      \multirow{2}{*}{SDD}           
        & $\mathbf{Img}$ & (80.0, 71.7) & (79.9\Same, 71.6\Same)
               & (85.5, 81.7) & (85.7\Same, 81.8\Same)
               & (81.4, 72.8) & (83.0\Up, 74.0\Up)
               & (87.3, 85.2) & (87.9\Up, 85.7\Up) \\
        & $\mathbf{Pix}$ & (79.8, 65.1) & (77.5\Down, 63.0\Down)
                & (90.7, 66.6) & (89.9\Down, 63.8\Down)
                & (71.7, 17.6) & (66.7\Down, 12.5\Down)
                & (88.7, 69.6) & (92.4\Up, 73.4\Up) \\
      \midrule
      \multirow{2}{*}{DAGM}          
        & $\mathbf{Img}$ & (94.6, 84.0) & (95.0\Same, 84.5\Same)
               & (96.1, 90.2) & (97.3\Up, 92.0\Up)
               & (97.3, 93.1) & (96.8\Same, 92.2\Same)
               & (97.1, 91.8) & (98.5\Up, 92.8\Up) \\
        & $\mathbf{Pix}$ & (82.4, 66.2) & (80.5\Down, 64.0\Down)
                & (95.6, 91.0) & (93.1\Down, 90.2\Down)
                & (91.1, 62.3) & (90.2\Down, 58.8\Down)
                & (95.8, 91.8) & (96.0\Same, 91.4\Same) \\
      \midrule
      \multirow{2}{*}{DTD-Synthetic} 
        & $\mathbf{Img}$ & (86.7, 95.2) & (85.0\Down, 93.5\Down)
               & (93.7, 97.3) & (93.2\Same, 96.9\Same)
               & (97.5, 98.2) & (97.5\Same, 98.4\Same)
               & (95.3, 98.0) & (97.7\Up, 98.4\Up) \\
        & $\mathbf{Pix}$ & (95.3, 86.9) & (93.5\Down, 84.5\Down)
                & (97.8, 91.1) & (93.3\Down, 83.4\Down)
                & (98.2, 86.5) & (94.9\Down, 73.3\Down)
                & (97.3, 89.9) & (98.0\Up, 90.2\Same) \\
      \bottomrule
    \end{tabular}
  \end{adjustbox}
\end{table}

\subsection{Comparison with State‑of‑the‑Art}
As reported in Table~\ref{table:industrial_results}, PILOT establishes state-of-the-art performance across all benchmarks for both image-level anomaly detection and pixel-level localization in the zero-shot setting. This can be benefited from PILOT comprehensive approach in addressing critical challenges prevalent in ZSAD, that explicitly tackles domain shift in test phase. Prior approaches~\cite{saa, winclip} rely solely on predefined prompts, which may not generalize well to unseen domains, whereas methods such as~\cite{April_GAN, AdaCLIP, CoCoOp, AnomalyCLIP} employ lightweight visual adapters or a single learnable prompt to capture anomaly cues, but often overfit to the auxiliary training set, limiting robustness under distribution shift. PILOT distinguishes itself by dynamically identifying the most relevant prompts for each input and refining only those learnable prompt parameters during inference using high-confidence pseudo-labels from the unlabeled target data. Note that while existing baselines can utilize the proposed image-level TTA strategies, these often degrade pixel-level localization performance and do not yield consistent improvements across both detection and localization tasks. On the contrary, PILOT uniquely achieves gains in both image-level and pixel-level metrics under TTA, a phenomenon analyzed in detail in Section~\ref{exp:TTA_effect}. We also compare against state‑of‑the‑art full‑shot methods in Appendix~F.1, which show that PILOT achieves image‑level accuracy on par.

\subsection{Effect of Test-Time Adaptation}
\label{exp:TTA_effect}

Table~\ref{table:tta_ab} presents a comparison of all methods evaluated under a unified TTA protocol, where each method adapts learnable parameters across the test set using high-confidence pseudo-labels following PILOT's label-free strategy. This fair setup reveals a key limitation of image-level TTA strategies in existing ZSAD models: while they can yield modest improvements in detection, they often destabilize pixel-level localization due to indiscriminate parameter updates guided by global objectives. This is particularly evident in methods like AnomalyCLIP, which use a single learnable prompt across all test images. In these cases, high-confidence pseudo-labels are often insufficient to represent the diversity of anomaly patterns in the target domain, resulting in imbalanced adaptation (see Appendix~F.2) and degraded fine-grained localization. Thus, image-level TTA alone struggles to maintain the alignment between visual and semantic cues necessary for precise localization, especially in complex scenarios. AAnother significant limitation is conventional TTA methods' susceptibility to noisy pseudo-labels, which propagate errors and degrade both detection and localization performance. PILOT’s adaptive prompt weighting addresses this by localizing the impact of label noise—updates concentrate on prompts closely aligned with each input, thus minimizing negative effects on the overall prompt pool.

\subsection{Module Ablation}
\label{experiment:Module_Ablation}
\begin{table}[h]
  \centering
  \scriptsize
\caption{
  Module ablation for PILOT on industrial datasets. 
  \textbf{SLP}: Single Learnable Prompt; 
  \textbf{+Attr}: SLP + Attribute Memory Bank; 
  \textbf{Full}: SLP + Attr + Query \& Prompt Pool (Full model).
  Metrics: Image = (AUROC, AP), Pixel = (AUROC, AUPRO).
}
  \label{table:module_ablation}
  \begin{tabular}{l|cc|cc}
    \toprule
    & \multicolumn{2}{c|}{\textbf{Image-Level}} & \multicolumn{2}{c}{\textbf{Pixel-Level}} \\
    \cmidrule(lr){2-3} \cmidrule(lr){4-5}
    & w/o TTA & w/ TTA & w/o TTA & w/ TTA \\
    \midrule
    SLP        & (90.8, 91.6) & (91.4\Up, 91.9\Same) & (94.2, 82.1) & (90.7\Down, 79.8\Down) \\
    +Attr      & (90.5, 91.3) & (91.2\Up, 91.7\Same) & (93.9, 81.5) & (94.0\Same, 81.5\Same) \\
    Full       & (90.3, 91.1) & (92.6\Up, 92.7\Up) & (93.5, 81.5) & (94.8\Up, 82.9\Up) \\
    \bottomrule
  \end{tabular}
\end{table}

\noindent Table~\ref{table:module_ablation} evaluates individual components of PILOT across industrial datasets, highlighting how our design stabilizes pixel-level predictions during TTA. Using a single prompt, similar to existing ZSAD methods, improves detection but severely degrades localization accuracy due to the lack of spatial supervision, causing visual-textual misalignment. Introducing the fixed attribute memory bank addresses this by providing stable semantic anchors, constraining model adaptation and preserving localization quality through parameter-free orthogonal projection. The complete PILOT framework integrates prompts from the learnable pool and attribute memory bank through a dual-branch fusion mechanism, dynamically weighted by adaptive query functions. During TTA, parameter updates are guided by alignment scores, ensuring prompts aligned with high-confidence samples undergo more substantial refinement. This adaptive update strategy mitigates negative transfer risks from unseen data, enhancing both image-level detection and pixel-level localization.

\section{Conclusion}
\label{sec:conclusion}

We have introduced PILOT, a ZSAD framework designed to address domain shifts by adaptively integrating anomaly cues through a dual-branch prompt aggregation mechanism, which combines a learnable prompt pool with semantic attribute embeddings. Our image-level TTA strategy further refines the prompt parameters using high-confidence pseudo-labels from unlabeled target data. Extensive experiments on 13 public benchmarks demonstrate that PILOT consistently achieves state-of-the-art performance in both anomaly detection and localization. Future work may explore advanced TTA strategies to jointly improve pixel-level localization and image-level detection, further enhancing ZSAD robustness.

\section*{Acknowledgements}
The authors wish to acknowledge the financial support of McGill University, the Natural Sciences and Engineering Research Council of Canada (NSERC), and the Fonds de Recherche du Québec – Nature et technologies (FRQNT). We also acknowledge the computing resources provided by Calcul Québec and the Digital Research Alliance of Canada.

\bibliography{1_Arvix}
\clearpage 

\appendix
\begin{center}
    \LARGE\bfseries APPENDIX
\end{center}
\vspace{0.5em}

This appendix provides supplementary details for the
BMVC 2025 paper titled ”Zero-Shot Anomaly Detection with Dual-Branch Prompt Selection”.

\begin{itemize}
  \item Appendix~\ref{sec:relatedwork} provides a comprehensive reviews of related works.
  \item Appendix~\ref{app:dataset} reviews all datasets.
  \item Appendix~\ref{app:baseline} provides information on the baseline methods.
  \item Appendix~\ref{app:implementation} details implementation specifics.
  \item Appendix~\ref{app:more_ablation} presents additional ablation studies for PILOT.
  \item Appendix~\ref{app:prompt_diversity_full} presents a cross-dataset analysis of prompt diversity and semantics.
  \item Appendix~\ref{app:Medical_localization} reports medical localization results without TTA.
  \item Appendix~\ref{app:Computational_Analysis} provides a computational analysis of PILOT compared to existing baselines.
  \item Appendix~\ref{app:comp_AnovL} compares AnovL with its TTA strategy.
  \item Appendix~\ref{app:attribute_bank_construction} describes the construction of the attribute memory bank.
  \item Appendix~\ref{app:Pseudo_code} provides the pseudo code for PILOT.
  \item Appendix~\ref{app:visualization} shows qualitative visualizations of anomaly localization.
\end{itemize}

\section{Related Work}
\label{sec:relatedwork}

\s{Prof Comment: USING an auxiliary dataset? yes mention it here as the sentence is not informative} \s{Prof Comment: HERE you need to have a connecting sentence as you have been talking about Auxilary-based datsets but suddenly switching to having no such a set just a bit tweaking would be nice to apply here} 
\s{Prof Comment: CAN you explain in short what is matched with what? you would like to have an informative sentence, the current one is not teaching the reader anything! }
\s{Prof Comment: INDICATE that these are for both image-level anomaly detection and pixel-level localization}
\noindent\textbf{Zero-shot Anomaly Detection.} ZSAD has gained increasing attention as a solution for identifying anomalies in unseen categories without requiring task-specific training data \cite{aota2023zero, winclip}. Leveraging the generalization capabilities of Vision–Language Models (VLMs) such as CLIP \cite{clip}, many approaches extend these models by linking image features to textual descriptions of normal and anomalous states. For instance, CLIP-AD \cite{clipad} and ZOC \cite{zoc} are early studies that leverage CLIP for ZSAD, but primarily focus on anomaly classification rather than localization. WinCLIP \cite{winclip} uses multiple handcrafted prompts with iterative patch-based forward passes, and April-GAN \cite{April_GAN} refines local semantics via fine-tuning on an auxiliary dataset; both methods address image-level anomaly detection and pixel-level localization. More recent prompt learning methods, such as AnomalyCLIP \cite{AnomalyCLIP}, AnomalyGPT \cite{anomalygpt}, and AdaCLIP \cite{AdaCLIP}, optimize object-agnostic text prompts in a single pass using auxiliary dataset. However, reliance on auxiliary datasets may introduce bias and compromise the intended zero-shot setting. This trade-off has been examined by Marzullo et al. \cite{marzullo2024exploring}, who observed that while avoiding auxiliary data reduces fine-tuning overhead, static prompts often fail to adapt to domain shifts; conversely, fine-tuning on auxiliary sets can improve accuracy but risks dataset bias and reduced generalization.
Our work aims at addressing these challenges by introducing a dual-branch prompt learning mechanism that combines learnable and fixed prompts, enabling more robust and adaptive anomaly detection under domain shift.

\noindent\textbf{CLIP Adaptation.} 
Recent advances in CLIP fine‑tuning primarily rely on prompt learning, in which learnable tokens are introduced into the text encoder to align visual features with related semantics~\cite{CoOp}. Approaches such as CoCoOp~\cite{CoCoOp} and visual prompt tuning~\cite{visualpromptting} condition the prompts on image content to improve robustness under domain shift. MaPLe~\cite{khattak2023maple} extends prompt learning by inserting learnable tokens into both the visual and text encoders of CLIP, jointly optimizing them to enhance cross-modal alignment. These techniques achieve parameter‑efficient adaptation by updating only the prompt representations and, in some cases, only CLIP’s final visual projection layer~\cite{fahes2024fine}. Following this rationale, our method fine‑tunes the visual projection layer to minimize additional parameters.

\s{Prof Comment: Similar entropy-minimization objectives and online normalization statistics updates ??? NOT CLEAR revise.}
\noindent\textbf{Test Time Adaptation.} A related direction for adapting CLIP-based models, particularly under domain shift, is TTA. In TTA, the model updates its parameters using unlabeled test data, often guided by pseudo-labels or self-supervised objectives. For instance, self-supervised losses based on the model’s own outputs can fine-tune representations without ground-truth labels~\cite{sun2020test}, and momentum-based pseudo-labeling helps enforce consistency across different augmented views~\cite{goyal2022test}. Other TTA strategies include entropy minimization, which guides the model to assign higher confidence to its predictions on unlabeled data, and the online updating of batch normalization statistics, which helps the model better align with the target distribution during test time. These approaches have been shown to improve robustness and generalization in image classification~\cite{nado2020evaluating} and segmentation~\cite{colomer2023adapt}. However, recent studies have found that these techniques are often less effective when the domain shift is large or when the model relies on strong vision-language priors, as in the case of CLIP. To our knowledge, AnovL~\cite{deng2023anovl} is the only TTA method for zero-shot anomaly localization, which employs an adapter module for pixel-level refinement and thus introduces additional parameters while focusing solely on localization. In contrast, our method adaptively updates the relevant learnable prompts, incurring no additional parameters and delivering consistent improvements in both image-level anomaly detection and pixel-level anomaly localization.
\section{Dataset}
\label{app:dataset}

\subsection{Benchmark Datasets}
\label{app:dataset_detail}

Appendix~\ref{app:dataset} provides a comprehensive review of the thirteen benchmark datasets utilized in our study, as summarized in Table~\ref{table:dataset_overview}. To rigorously evaluate both image-level detection and pixel-level localization capabilities of PILOT, we select six industrial inspection datasets. DAGM~\cite{dataset_DAGM} and DTD-Synthetic~\cite{dataset_DTD} comprise textured surfaces exhibiting subtle variations in material appearance, presenting unique challenges for texture-based anomaly detection due to the fine-grained and ambiguous nature of defects. In contrast, MVTec AD~\cite{dataset_mvtec}, VisA~\cite{dataset_visa}, BTAD~\cite{dataset_btad}, and SDD~\cite{dataset_ksdd} focus on object-centric industrial scenarios, with defects such as cracks, scratches, dents, and missing components distributed across everyday manufactured and synthetic objects. These datasets are characterized by diverse object categories and high variability in anomaly manifestation, supporting comprehensive benchmarking of localization and generalization performance.

\begin{table}[htbp]
\centering
\caption{
Summary of the 13 public benchmark datasets. ``Ind'' = Industrial, ``Med'' = Medical, ``Def'' = Defect detection, ``AD'' = Anomaly detection, ``AL'' = Anomaly localization.
}
\label{table:dataset_overview}
\begin{tabular}{@{}lcccccl@{}}
\toprule
\textbf{Dataset} & \textbf{Cls.} & \textbf{Norm} & \textbf{Anom} & \textbf{Total} & \textbf{Dom.} & \textbf{Task} \\
\midrule
MVTec AD      & 15 & 467  & 1258 & 1725 & Ind & Def \\
VisA          & 12 & 962  & 1200 & 2162 & Ind & Def \\
BTAD          &  3 & 451  &  290 &  741 & Ind & Def \\
SDD           &  1 & 181  &   74 &  255 & Ind & Def \\
DAGM          & 10 & 6996 & 1054 & 8050 & Ind & Def \\
DTD‑Synthetic & 12 & 357  &  947 & 1304 & Ind & Def \\
\midrule
HeadCT        &  1 & 100  &  100 &  200 & Med & AD \\
BrainMRI      &  1 &  98  &  155 &  253 & Med & AD \\
Br35H         &  1 & 1500 & 1500 & 3000 & Med & AD \\
ISIC          &  1 &   0  &  379 &  379 & Med & AL \\
CVC‑ClinicDB  &  1 &   0  &  612 &  612 & Med & AL \\
CVC‑ColonDB   &  1 &   0  &  380 &  380 & Med & AL \\
Kvasir        &  1 &   0  & 1000 & 1000 & Med & AL \\
\bottomrule
\end{tabular}
\end{table}

To assess clinical applicability, we include seven medical imaging benchmarks. Three brain scan datasets, HeadCT~\cite{dataset_headCT}, BrainMRI~\cite{dataset_brain}, and Br35H~\cite{dataset_br35h}, comprise both healthy and pathological subjects, facilitating binary anomaly classification tasks on volumetric neuroimaging data. These datasets encompass a spectrum of anatomical and pathological variations, thereby enabling evaluation of robustness to inter-patient heterogeneity and rare conditions. The remaining four datasets provide pixel-wise lesion annotations and consist exclusively of abnormal samples. ISIC~\cite{dataset_isic} features dermoscopic images annotated for skin lesions, while CVC-ClinicDB~\cite{dataset_ClinicDB}, CVC-ColonDB~\cite{dataset_ColonDB}, and Kvasir~\cite{dataset_Kvasir} contain endoscopic images of the gastrointestinal tract, highlighting a range of polyp and ulcer appearances. Since these medical localization datasets lack normal examples, test-time adaptation is not performed for them; their localization results without TTA are reported in Appendix~\ref{app:Medical_localization}.

\subsection{Domain Discrepancy Analysis}
\label{app:domain_gap}

To validate that MVTec AD and VisA indeed represent statistically distinct domains suitable for strict zero-shot evaluation, we quantitatively assessed the distributional discrepancy between their CLIP global feature embeddings. Specifically, we employed Maximum Mean Discrepancy (MMD$^2$)~\cite{gretton2012kernel}, Fréchet Inception Distance (FID)~\cite{heusel2017gans}, and the Kolmogorov--Smirnov (KS)~\cite{massey1951kolmogorov} test. Our analysis yielded $\mathrm{MMD}^2 = 0.0328$ ($p$-value~$= 0.0099$), indicating a significant difference in embedding distributions. The FID score between the two datasets is $51.9$, further corroborating a substantial distributional gap. Additionally, the KS test computed across all feature dimensions revealed that the minimum, mean, and maximum $p$-values are $0$, $5.16 \times 10^{-5}$, and $0.0329$, respectively, suggesting statistically significant differences in nearly every dimension of the embedding space. These results collectively confirm that MVTec AD and VisA are non-overlapping in the learned representation space, thus justifying their use as auxiliary and target datasets in our zero-shot anomaly detection experiments.

\section{Baseline}
\label{app:baseline}

In Appendix~\ref{app:baseline}, we compare PILOT against six representative zero‑shot anomaly detection methods, summarized as follows:

\begin{itemize}
  \item \textbf{SAA} \cite{saa}: Relies on a pretrained Grounding DINO to propose candidate anomaly regions, which are then segmented by the Segment Anything Model (SAM) without any additional training.
  
  \item \textbf{CoCoOp} \cite{CoCoOp}: Replaces fixed CLIP prompt tokens with learnable context vectors while keeping the backbone frozen. Offers both a shared “unified” prompt and class‑specific prompts to adapt flexibly to unseen categories.
  
  \item \textbf{WinCLIP} \cite{winclip}: Splits the image into overlapping windows, encodes each via CLIP to produce localized anomaly scores, and aggregates window predictions through an ensemble of token embeddings—improving spatial localization without fine‑tuning.
  
  \item \textbf{APRIL‑GAN} \cite{April_GAN}: Builds on WinCLIP by refining text prompts for tighter vision–language alignment and injecting lightweight learnable projections into the visual encoder, thereby sharpening anomaly masks while retaining zero‑shot generality.
  
  \item \textbf{AnomalyCLIP} \cite{AnomalyCLIP}: Learns object‑agnostic “normal” and “anomalous” text prompts, then employs a dense patch‑to‑prompt attention module (DPAM) to align each visual patch embedding with the appropriate prompt for pixel‑level localization.
  
  \item \textbf{AdaCLIP} \cite{AdaCLIP}: Combines static prompts shared across all inputs for coarse adaptation with dynamic, image‑conditioned prompts for fine‑grained adjustment. Both prompt sets are jointly learned on auxiliary anomaly data to balance detection accuracy and localization precision.
\end{itemize}

\section{Implementation Details}
\label{app:implementation}

\subsection{PILOT Details}
\label{app:PILOT_Details}

Appendix~\ref{app:implementation} provides full implementation details of PILOT. We implement PILOT atop the open-source CLIP ViT-L/14@336px backbone, freezing all transformer weights except for the final visual-to-joint projection layer, which is jointly optimized alongside our prompt modules. Our learnable prompt pool $\mathcal{P}$ contains $K=5$ paired prompts that each comprise $L=12$ learnable vectors for ``normal'' and ``anomalous'' semantics, while a fixed attribute memory bank (see Appendix~\ref{app:attribute_bank_construction}) provides semantic tokens extracted from structured templates. All images are uniformly resized to $518\times518$ pixels, and we extract both the [CLS] token \(x_{\mathrm{cls}}\) and patch embeddings \(\{X^{(M)}\}\) at layers 6, 12, 18, and 24 of the CLIP encoder. All prompt embeddings are initialized by sampling from a Gaussian distribution $\mathcal{N}(0,\,0.02)$. PILOT is trained for five epochs using AdamW with a base learning rate of $10^{-3}$, a batch size of 32. During test-time adaptation, we fix the pseudo-label fraction at $\rho=0.25$. All experiments are conducted in PyTorch~2.1.0 on a single NVIDIA Ada A5000 GPU with 32~GB of memory.

\subsection{Anomaly State Descriptors}
\label{app:anomaly_state_descriptors}

The following anomaly state descriptors are used in our nearest-neighbor analysis, as described in Appendix~\ref{app:prompt_diversity_full} of the main text. These descriptors were selected to cover a wide range of mechanical, surface, electrical, pressure-related, assembly, and contamination defects:
\begin{itemize}
  \item \textbf{Mechanical damage:} damaged, scratched, cracked, chipped, broken, bent, punctured, fractured, delaminated, weld defect, loose component, misaligned, deformed, twisted, buckled
  \item \textbf{Surface \& corrosion:} rusty, corroded, oxidized, pitted, eroded, paint peeling, stained, contaminated, oil leak, grease smear
  \item \textbf{Thermal \& electrical:} overheated, burnt, scorched, melted, short-circuited, electrical arcing, voltage spike damage
  \item \textbf{Sealing \& pressure:} leaking, sealing failure, gasket blown, under-pressure, over-pressure, gas leak
  \item \textbf{Wear \& fatigue:} worn bearing, surface fatigue crack, abrasion mark, material fatigue, delamination
  \item \textbf{Assembly \& alignment:} loose screw, missing bolt, stripped thread, cross-threaded, flange misfit
  \item \textbf{Contamination \& blockages:} clogged, blocked channel, foreign particle inclusion, scale buildup, slag inclusion
  \item \textbf{Other:} vibration damage, impact dent, surface blistering, vacuum leak, structural crack
\end{itemize}

\subsection{Entmax\texorpdfstring{$_{1.5}$}{₁.₅} Details}
\label{app:entmax}

Let $\boldsymbol{\kappa}=(\kappa_{1},\dots,\kappa_{D})\in\mathbb{R}^{D}$ be a vector of input scores, where each $\kappa_{i}\in\mathbb{R}$ denotes the unnormalized score for the $i$‑th dimension of interest, and write the probability simplex as
\begin{equation}
\Delta^{D}
=
\bigl\{\boldsymbol{\chi}\in\mathbb{R}^{D}\mid \chi_{i}\ge 0,\;\sum_{i=1}^{D}\chi_{i}=1\bigr\},
\end{equation}
where $\Delta^{D}$ is the set of all $D$‑dimensional probability distributions.

The $\mathrm{entmax}_{1.5}$ operator produces a sparse probability vector $\boldsymbol{\chi}\in\Delta^{D}$ via
\begin{equation}
\chi_{i}
=
\bigl[\max\bigl(0,\;\kappa_{i}-\tau(\boldsymbol{\kappa})\bigr)\bigr]^{2},
\quad
i=1,\dots,D,
\end{equation}
where the scalar $\tau(\boldsymbol{\kappa})\in\mathbb{R}$ is chosen so that $\sum_{i=1}^D\chi_{i}=1$.  Intuitively, any entry with $\kappa_{i}\le\tau(\boldsymbol{\kappa})$ is set to zero, and the remaining values are shifted, squared, and renormalized.  This yields a differentiable, piecewise‑quadratic mapping that retains only the most salient dimensions in its support. In PILOT, injecting Gumbel noise before applying entmax$_{1.5}$ ensures that only the highest‑scoring attribute cues survive to influence the final text embedding, providing robust, sparse weighting during training.

\s{Shouldn't below be a separate subsection? -- yes, agree. I will make this into subsection   It is not the right place? as you are talking experiments. this is more experimental details.... am i right? -- right, i will put it into different subsection   try to put all these kind of implementation details first and then the result reletaded sections... similar to what we do in the main document... -- okay, i will}

\section{More Ablation Study}
\label{app:more_ablation}

\subsection{Comparison with SOTA full-shot method}
\label{app:full_shot}
\s{Prof Comment: The caption is too long! summarize it. -- okay   so I see we have reds in the caption that is why it looks too long  but see if you can summarize -- yes, i will,     Have you  discussed in the text that what do you mean by Full-shot? -- no i will add the definition of full shot, sorry i assume people know it again...}
\begin{table}[h]
    \centering
    \caption{Performance comparison on industrial anomaly detection benchmarks. PILOT is evaluated in the zero-shot setting; all other methods are full-shot. Top section: image-level (AUROC, AP); bottom: pixel-level (AUROC, AUPRO). Best results in bold, second-best underlined.}
    \label{table:full_shot}
    \begin{tabular}{@{}lcccc@{}}
        \toprule
        \multicolumn{5}{c}{\textbf{Industrial Image‑level Anomaly Detection (AUROC, AP)}} \\
        \midrule
        Dataset        & PILOT               & PatchCore                     & CDO                             & RD4AD                          \\
        \midrule
        MVTec AD       & (92.1, 96.6)        & (\textbf{99.0}, \textbf{99.7})   & (97.1, 97.2)                     & (\underline{98.7}, \underline{99.4})   \\[0.5ex]
        VisA           & (84.1, 85.7)        & (94.6, \textbf{95.9})            & (\underline{95.0}, 95.1)         & (\textbf{95.3}, \underline{95.7})     \\[0.5ex]
        BTAD           & (\underline{95.3}, 96.8) & (93.2, \textbf{98.6})         & (\textbf{97.6}, \underline{97.7}) & (93.8, 96.8)                        \\[0.5ex]
        SDD            & (87.9, 85.7)        & (\underline{91.8}, \underline{91.6}) & (\textbf{96.0}, \textbf{96.2}) & (86.8, 81.3)                        \\[0.5ex]
        DAGM           & (\textbf{98.5}, \underline{92.8}) & (95.4, 90.3)      & (\underline{95.1}, \textbf{95.0}) & (92.9, 79.1)                        \\[0.5ex]
        DTD‑Synthetic  & (\textbf{97.7}, \textbf{98.4}) & (\underline{97.5}, \underline{97.3}) & (96.8, 96.9)             & (96.2, 96.3)                        \\[0.5ex]
        \midrule
        \multicolumn{5}{c}{\textbf{Industrial Pixel‑level Anomaly Localization (AUROC, AUPRO)}} \\
        \midrule
        Dataset        & PILOT               & PatchCore                     & CDO                             & RD4AD                          \\
        \midrule
        MVTec AD       & (90.3, 80.2)        & (\underline{98.1}, 92.8)       & (\textbf{98.2}, \textbf{98.1})    & (97.8, \underline{97.6})           \\[0.5ex]
        VisA           & (96.0, 87.2)        & (98.5, 92.2)                   & (\textbf{99.0}, \textbf{99.2})    & (\underline{98.4}, \underline{98.9}) \\[0.5ex]
        BTAD           & (96.3, 74.9)        & (97.4, 74.4)                   & (\textbf{98.1}, \textbf{98.2})    & (\underline{97.5}, \underline{97.8}) \\[0.5ex]
        SDD            & (\underline{92.4}, 73.4) & (87.9, 46.3)              & (\textbf{97.9}, \textbf{97.7})    & (92.2, \underline{92.0})           \\[0.5ex]
        DAGM           & (96.0, 91.4)        & (95.9, 87.9)                   & (\textbf{97.3}, \textbf{97.2})    & (\underline{96.8}, \underline{96.9}) \\[0.5ex]
        DTD‑Synthetic  & (98.0, 90.2)        & (\underline{98.2}, \underline{98.1}) & (\textbf{98.3}, \textbf{98.5}) & (97.6, 97.7)                        \\[0.5ex]
        \bottomrule
    \end{tabular}
\end{table}

In this comparison, we evaluate PILOT alongside three state-of-the-art fully supervised methods—PatchCore~\cite{roth2022towards}, CDO~\cite{cao2023collaborative}, and RD4AD~\cite{deng2022anomaly}—across six widely used industrial benchmarks. Here, a full-shot method refers to a model trained with full access to samples from the target domain. While these baselines leverage direct access to target domain knowledge, PILOT operates solely with pretrained vision–language representations and requires no information from the target domain. Remarkably, PILOT achieves detection accuracy on par with these full-shot models, even when faced with unseen textures or novel defect types. It matches PatchCore’s memory-based robustness, rivals the collaborative feature learning of CDO, and maintains high sensitivity to both global and local irregularities, comparable to RD4AD’s reconstruction-driven approach. However, there remains room for improvement in pixel-level localization accuracy, which we identify as a promising direction for future research. These results underscore the significant potential of zero-shot anomaly detection frameworks in real-world industrial scenarios.

\subsection{Pseudo label Statistic}

\label{app:pseudo_stats}
\begin{table*}[h]
    \centering
    \caption{TTA Dataset selection distribution statistics across datasets under different pseudo‑labeling ratios \(\rho\). M1: Shannon Evenness, M2: Coefficient of Variation, M3: Percentage of Noisy Label.}
    \label{table:tta_distribution_metrics}
    \begin{adjustbox}{width=\textwidth}
    \begin{tabular}{@{}l c *{5}{ccc}@{}}
        \toprule
        Dataset    & Class 
                   & \multicolumn{3}{c}{$\rho=$5\%}
                   & \multicolumn{3}{c}{$\rho=$15\%}
                   & \multicolumn{3}{c}{$\rho=$25\%}
                   & \multicolumn{3}{c}{$\rho=$35\%}
                   & \multicolumn{3}{c}{$\rho=$50\%} \\
        \cmidrule(lr){3-5} \cmidrule(lr){6-8} \cmidrule(lr){9-11} \cmidrule(lr){12-14} \cmidrule(lr){15-17}
                   &        
                   & M1   & M2   & M3  
                   & M1   & M2   & M3  
                   & M1   & M2   & M3  
                   & M1   & M2   & M3  
                   & M1   & M2   & M3  \\
        \midrule
        MVTec AD      &   15   & 0.76  & 1.21  & 10.3\%  & 0.97  & 0.45  & 8.1\%   & 0.93  & 0.71  & 7.8\%   & 0.95  & 0.59  & 11.2\% & 1.00  & 0.00  & 27.3\% \\
        VisA          &   12   & 0.79  & 0.97  & 12.0\%  & 0.98  & 0.27  & 13.2\%  & 0.98  & 0.35  & 14.2\%  & 0.99  & 0.31  & 16.1\%  & 1.00  & 0.00  & 31.4\% \\
        BTAD          &   3   & 0.79  & 0.76  & 21.6\%  & 0.86  & 0.70  & 12.6\%  & 0.93  & 0.50  & 17.3\%  & 0.95  & 0.39  & 20.4\%  & 1.00  & 0.00  & 36.5\% \\
        SDD           &   1   &  0.72    &  0.59    & 0.0\%   &  0.61    &  0.69    & 7.6\%   &  0.63    &  0.68    & 6.3\%   &  0.74    &  0.58    & 10.1\%  &  1.00    &  0.00    & 39.6\% \\
        DAGM          &   10   & 0.73  & 1.35  & 3.9\%   & 0.88  & 0.76  & 1.3\%   & 0.95  & 0.56  & 0.7\%   & 0.97  & 0.43  & 0.9\%   & 1.00  & 0.00  & 6.9\%  \\
        DTD‑Synthetic &   12   & 0.72  & 1.34  & 3.0\%   & 0.96  & 0.47  & 6.1\%   & 0.95  & 0.53  & 4.6\%   & 0.96  & 0.48  & 5.2\%   & 1.00  & 0.00  & 20.7\% \\
        \bottomrule
    \end{tabular}
    \end{adjustbox}
\end{table*}

\s{In the above table, first two, write that $\rho=5\%$ currently you only have written the numbers --- $\rho$ is not written in the table.... do you get it...  the first row is $\rho$ but it is not clerely defined unless i look at the caption and implicitly understand that yes the first is that.... -- i see, the $\rho$ is defined in method, i will no new definition i needed. just write see my pointyer}
\s{you may define M1-3 here in the text when talking about your three metrics. -- okay}
In Table~\ref{table:tta_distribution_metrics} we report three statistics that describe how our TTA selection (See Section~\ref{method:TTA} of the main paper) distributes pseudo‑labels over the target dataset~\(D_{\mathrm{tar}}\). Let \(B = \lvert D_{\mathrm{tar}}\rvert\) and recall that we form the top‑\(\rho\) set \(A\) and bottom‑\(\rho\) set \(N\), with \(S = A \cup N\). For each class \(i\) in the set of classes \(\mathcal I\), we write \(g_i\) for the fraction of samples in \(S\) belonging to class \(i\).

\s{is the blue correct? -- yes write a similar sentence for the other 2 metrics. -- Okay! }
\begin{itemize}
  \item \textbf{Shannon evenness} \(E\) is defined as
    \[
      E = -\frac{\sum_{i\in\mathcal I} g_i \ln g_i}{\ln \lvert\mathcal I\rvert}.
    \]
    Values near 1 indicate that the selected samples \(S\) are spread uniformly across all classes. This is referred as M1 in Table~\ref{table:tta_distribution_metrics}.

  \item \textbf{Coefficient of variation} \(\mathrm{CV}\) is
    \[
      \mathrm{CV} = \frac{\sigma(g)}{\mu(g)},
    \]
    where \(\mu(g)\) and \(\sigma(g)\) are the mean and standard deviation of \(\{g_i\}\). A low \(\mathrm{CV}\) signals a balanced class‑wise count. This is referred as M2 in Table~\ref{table:tta_distribution_metrics}.

  \item \textbf{Noisy label rate} is the fraction of indices in \(S\) whose pseudo‑label \(\tilde y_j\) disagrees with the ground‑truth label \(y_j\). This is referred as M3 in Table~\ref{table:tta_distribution_metrics}.
\end{itemize}

As \(\rho\) grows, we observe that \(E\) increases and \(\mathrm{CV}\) decreases, confirming that selecting a larger fraction of samples both diversifies class coverage and balances the counts. At the same time, the noisy label rate rises, reflecting the classic trade‑off: higher coverage under TTA comes at the expense of pseudo‑label accuracy.

\subsection{Other TTA Method}
\label{app:tta_ablation}

To better understand the impact of different TTA strategies on anomaly detection performance, we conduct an ablation study comparing several representative TTA methods. As shown in Table~\ref{table:tta_method_ablation}, we evaluate TENT~\cite{wang2020tent}, BN Adaptation~\cite{lim2023ttn}, and soft pseudo labeling alongside our proposed PILOT framework.

\begin{table}[h]
    \centering
    \caption{Ablation of different TTA methods on industrial anomaly detection datasets. Metrics reported: Image-level (AUROC, AP) and Pixel-level (AUROC, AUPRO).}
    \label{table:tta_method_ablation}
    \begin{tabular}{@{}l|cc|cc@{}}
        \toprule
        \multirow{2}{*}{\textbf{TTA Method}} 
            & \multicolumn{2}{c|}{\textbf{Image-level}} 
            & \multicolumn{2}{c}{\textbf{Pixel-level}} \\
        \cmidrule(lr){2-3} \cmidrule(lr){4-5}
            & AUROC & AP & AUROC & AUPRO \\
        \midrule
        TENT &  74.3   &  73.5   &  61.6   &  40.9  \\
        BN Adaptation &  70.5   &  69.8   &  55.2   &  34.5   \\
        Soft Pseudo Label    &  91.0   &  91.2   &  92.0   &  79.8   \\
        \midrule
        PILOT         &  92.6   &  92.7   &  94.8   &  82.9   \\
        \bottomrule
    \end{tabular}
\end{table}

Table~\ref{table:tta_method_ablation} presents the results of different TTA strategies on industrial anomaly detection benchmarks. Both TENT and BN Adaptation exhibit weak performance in this setting, particularly for pixel-level localization, suggesting that these conventional TTA methods are not well suited for adapting CLIP-based models. Recent literature has also observed that methods such as TENT, which adapt only batch normalization parameters, often struggle when applied to models with large distribution shifts or strong vision-language priors, as is the case with CLIP~\cite{liang2025comprehensive}. In contrast, the soft pseudo labeling strategy leads to more stable improvements, although its cautious approach may be less aggressive in leveraging adaptation signals. Our findings suggest that confidence-based pseudo-labeling, as used in the PILOT TTA framework, provides the most effective adaptation and yields the best overall performance in this scenario.

\subsection{Effect of Learnable Prompt Pool Size} 

\begin{table}[h]
  \centering
  \scriptsize
  \setlength{\tabcolsep}{3pt}
  \caption{Effect of prompt pool size $K$ on the industrial benchmark. Best in each row is \textbf{bold}.}
  \label{table:prompt_pool_size}
  \begin{tabular}{ll *{9}{c}}
    \toprule
    \multirow{2}{*}{\textbf{Level}}
      & \multirow{2}{*}{\textbf{Metric}}
      & \multicolumn{9}{c}{$K$} \\
    \cmidrule(lr){3-11}
      & & 2 & 3 & 4 & \textbf{5} & 7 & 10 & 16 & 32 & 64 \\
    \midrule
    \multirow{2}{*}{Image-level}
      & \textsc{AUROC} & 85.9 & 88.6 & 91.3 & 92.6 & 92.4 & 92.4 & 92.9 & \textbf{93.2} & 91.7 \\
      & \textsc{AP}    & 87.2 & 90.8 & 91.9 & 92.7 & 92.0 & 92.0 & 93.5 & \textbf{93.8} & 92.2 \\
    \addlinespace[2pt]
    \multirow{2}{*}{Pixel-level}
      & \textsc{AUROC} & 94.1 & 93.7 & 94.6 & \textbf{94.8} & 91.7 & 88.0 & 89.4 & 87.1 & 83.7 \\
      & \textsc{AUPRO} & 82.1 & 81.0 & 82.4 & \textbf{82.9} & 80.9 & 77.2 & 78.3 & 76.3 & 73.8 \\
    \bottomrule
  \end{tabular}
\end{table}

Table~\ref{table:prompt_pool_size} examines how changing the number of learnable prompts influences performance. As the pool size increases, image-level metrics rise steadily and then level off once a moderate number of prompts is reached, indicating improved anomaly discrimination with greater prompt diversity. However, pixel-level metrics follow a different trend: they increase up to a moderate pool size but then decline with further increases, reflecting reduced generalization for localization. This divergence shows that while a larger pool improves global detection, overly large pools lead to redundancy and overfitting at the pixel level, making a moderate pool size optimal for balancing both objectives.

\subsection{Effect of Attribute Memory Bank Size.} 

\begin{table}[h]
  \centering
  \scriptsize
  \setlength{\tabcolsep}{3pt}
  \caption{Effect of attribute-memory bank size $C$ on the industrial benchmark. Best in each row is \textbf{bold}.}
  \label{table:attr_bank_size}
  \begin{tabular}{ll *{6}{c}}
    \toprule
    \multirow{2}{*}{\textbf{Level}}
      & \multirow{2}{*}{\textbf{Metric}}
      & \multicolumn{6}{c}{$C$} \\
    \cmidrule(lr){3-8}
      & & 5 & 25 & 50 & \textbf{75} & 100 & 150 \\
    \midrule
    \multirow{2}{*}{Image-level}
      & \textsc{AUROC} & 88.1 & 92.1 & 92.5 & \textbf{92.6} & 92.7 & 92.2 \\
      & \textsc{AP}    & 88.4 & 92.0 & 92.4 & \textbf{92.7} & 92.6 & 92.3 \\
    \addlinespace[2pt]
    \multirow{2}{*}{Pixel-level}
      & \textsc{AUROC} & 87.7 & 93.9 & 94.5 & \textbf{94.8} & 94.6 & 94.4 \\
      & \textsc{AUPRO} & 76.1 & 82.2 & 82.7 & \textbf{82.9} & 82.8 & 82.5 \\
    \bottomrule
  \end{tabular}
\end{table}

\label{app:attribute_memory_bank_size}
Table~\ref{table:attr_bank_size} shows the impact of varying the number of attribute templates. Expanding the memory bank from a small to a moderate size leads to clear improvements in both image-level and pixel-level performance, as the model benefits from richer semantic information. Beyond this range, image-level metrics stabilize, while pixel-level gains plateau and may slightly decrease with the largest banks. This suggests that adding more attributes beyond a certain point does not further improve generalization, and that a compact but diverse attribute bank is sufficient to provide strong semantic guidance.

\subsection{Effect of $\rho$ fraction} 

\begin{table}[h]
  \centering
  \scriptsize
  \setlength{\tabcolsep}{4pt}
  \caption{Effect of pseudo-label fraction $\rho$ (\%) during TTA (industrial datasets).
           Best value in each column is \textbf{bold}.}
  \label{table:rho_fraction}
  \begin{tabular}{llccccc}
    \toprule
    \multirow{2}{*}{\textbf{Level}}
      & \multirow{2}{*}{\textbf{Metric}}
      & \multicolumn{5}{c}{$\rho$ (\%)}
      \\
    \cmidrule(lr){3-7}
      & & 5 & 15 & \textbf{25} & 35 & 50 \\
    \midrule
    \multirow{2}{*}{Image-level}
      & \textsc{AUROC} & 92.4 & 93.0 & \textbf{92.6} & 90.1 & 88.5 \\
      & \textsc{AP}    & 92.5 & 93.1 & \textbf{92.7} & 90.2 & 88.7 \\
    \addlinespace[2pt]
    \multirow{2}{*}{Pixel-level}
      & \textsc{AUROC} & 92.4 & 93.6 & \textbf{94.8} & 92.7 & 91.5 \\
      & \textsc{AUPRO} & 81.0 & 81.9 & \textbf{82.9} & 79.9 & 78.0 \\
    \bottomrule
  \end{tabular}
\end{table}

We vary the pseudo-label fraction $\rho$ during TTA, as shown in Table~\ref{table:rho_fraction}. Using a small set of high-confidence pseudo labels leads to improvements in image-level performance compared to no adaptation, as the adaptation process focuses on the most reliable samples. Notably, pixel-level performance remains largely unchanged when only a small subset is used, but begins to improve as $\rho$ increases to an intermediate fraction. At this optimal point, the model is able to leverage a greater diversity of anomaly patterns without introducing substantial label noise. Beyond this, further increases in $\rho$ admit less reliable pseudo labels, resulting in a sharp decline in both detection and localization metrics.
\subsection{Prompt Aggregation Ablation}
\label{app:prompt_aggregation_ablation}

\begin{table}[h]
  \centering
  \caption{Ablation for prompt aggregation. ``w/o TTA'' and ``w/ TTA'' indicate performance without and with TTA, respectively. ``Extra'' indicates whether additional trainable parameters are used. OP: orthogonal projector as used in PILOT; GF: Gate Fusion~\cite{arevalo2017gated}; CA: Cross Attention~\cite{chen2021crossvit}.}
  \label{table:Projector_ablation}
  \begin{adjustbox}{width=0.6\columnwidth,center}
    \begin{tabular}{@{}l c | cc | cc@{}}
      \toprule
      \textbf{Proj.} & \textbf{Extra} 
      & \multicolumn{2}{c}{\textbf{Image}} 
      & \multicolumn{2}{c}{\textbf{Pixel}} \\
      \cmidrule(lr){3-4} \cmidrule(lr){5-6}
      & & w/o TTA & w/ TTA & w/o TTA & w/ TTA \\
      \midrule
      GF & \(\checkmark\) 
        & (91.2, 90.0) & (91.5\Same, 90.8\Up) 
        & (88.6, 77.8) & (89.4\Up, 77.6\Same) \\
      CA & \(\checkmark\) 
        & (92.1, 92.4) & (93.1\Up, 92.9\Up) 
        & (93.7, 81.3) & (86.4\Down, 75.3\Down) \\
      OP & \(\times\) 
        & (90.3, 91.1) & (92.6\Up, 92.7\Up)
        & (93.5, 81.5) & (94.8\Up, 82.9\Up) \\
      \bottomrule
    \end{tabular}
  \end{adjustbox}
\end{table}

Table~\ref{table:Projector_ablation} compares three prompt aggregation strategies and reveals that methods like Gate Fusion~\cite{arevalo2017gated} and Cross Attention~\cite{chen2021crossvit}, while capable of improving image-level anomaly detection, tend to destabilize pixel-level localization during TTA. While these approaches may leverage the attribute memory bank as a semantic anchor, their extra parameters are directly updated using noisy pseudo-labels and only image-level supervision. This makes them prone to parameter drift, which can erode the stability needed for precise localization under domain shift. In contrast, the orthogonal projector used in PILOT achieves reliable gains across both detection and localization tasks without introducing extra parameters, as it keeps the attribute embedding fixed as a semantic reference and allows only the prompt representation to adapt. This encourages adaptation in directions that are truly complementary, while avoiding redundancy and overfitting.

\subsection{Effect of Orthogonal Projection}
\label{app:orthogonal_projection_tracker}
\s{caption is too long, keep the important parts and summarize. don't have formula in the caption, you may instead refer to the eq. nub,er e.g. here (20)-- okay, that would be great, i will revise it }
\begin{figure}[htbp]
  \centering
  \includegraphics[width=0.75\columnwidth]{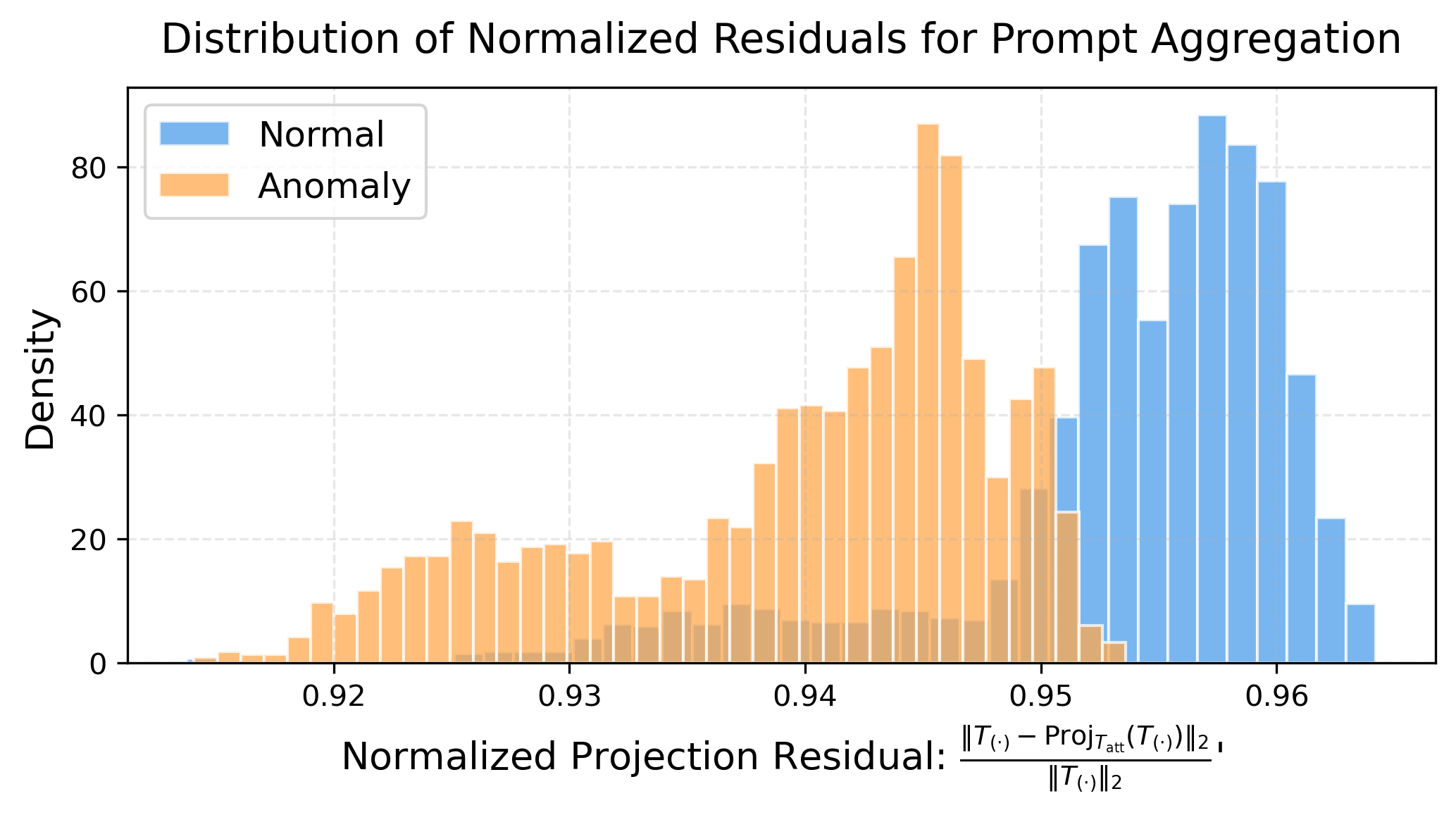}
  \caption{Histogram of normalized projection residuals for normal and anomaly prompts (see Eq.~\ref{eq:fuse-main}). The y-axis shows the fraction of samples per bin. Higher values indicate greater orthogonality to the anchor, while lower values reflect stronger alignment.}
  \label{fig:projection_visual}
\end{figure}

To evaluate the effectiveness of our orthogonal projection strategy for prompt aggregation (Equation~\ref{eq:fuse-main}), we analyze the distribution of normalized projection residuals defined as:
\begin{equation}
\frac{\| T_{(\cdot)} - \mathrm{Proj}_{T_{\mathrm{att}}}(T_{(\cdot)}) \|_2}{\| T_{(\cdot)} \|_2}
\end{equation}
for both learnable prompts $T_n$ (normal) and $T_a$ (anomaly), as shown in Figure~\ref{fig:projection_visual}. The resulting histograms exhibit consistently nonzero residuals, which are clearly separated between normal and anomalous samples. This indicates that the learnable prompts encode substantial information complementary to the attribute embedding $T_{\mathrm{att}}$, rather than merely duplicating its semantics. Notably, anomalous prompts display larger orthogonal components, highlighting the model's ability to leverage diverse information captured by the learnable prompts. Empirically, these findings confirm that the orthogonal projection mechanism extracts and integrates useful, non-redundant cues into the fused embedding, thereby validating our parameter-free aggregation approach.

\section{Prompt Diversity}
\label{app:prompt_diversity_full}

\begin{figure}[htbp]
  \centering
  \subfigure[Prompt-pair similarity]{\includegraphics[width=0.48\columnwidth]{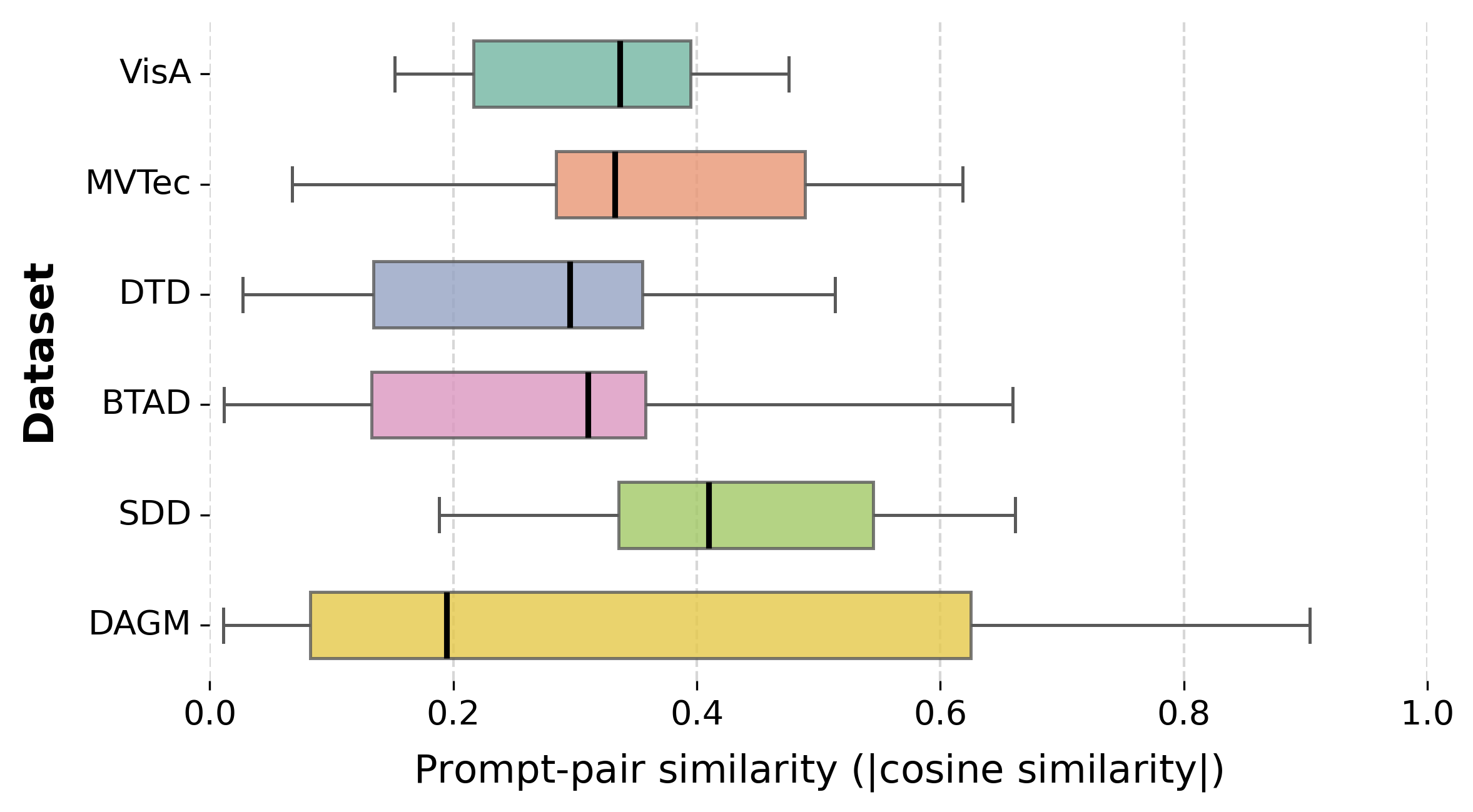}\label{fig:prompt_diversity}}
  \hfill
  \subfigure[Prompt contribution]{\includegraphics[width=0.48\columnwidth]{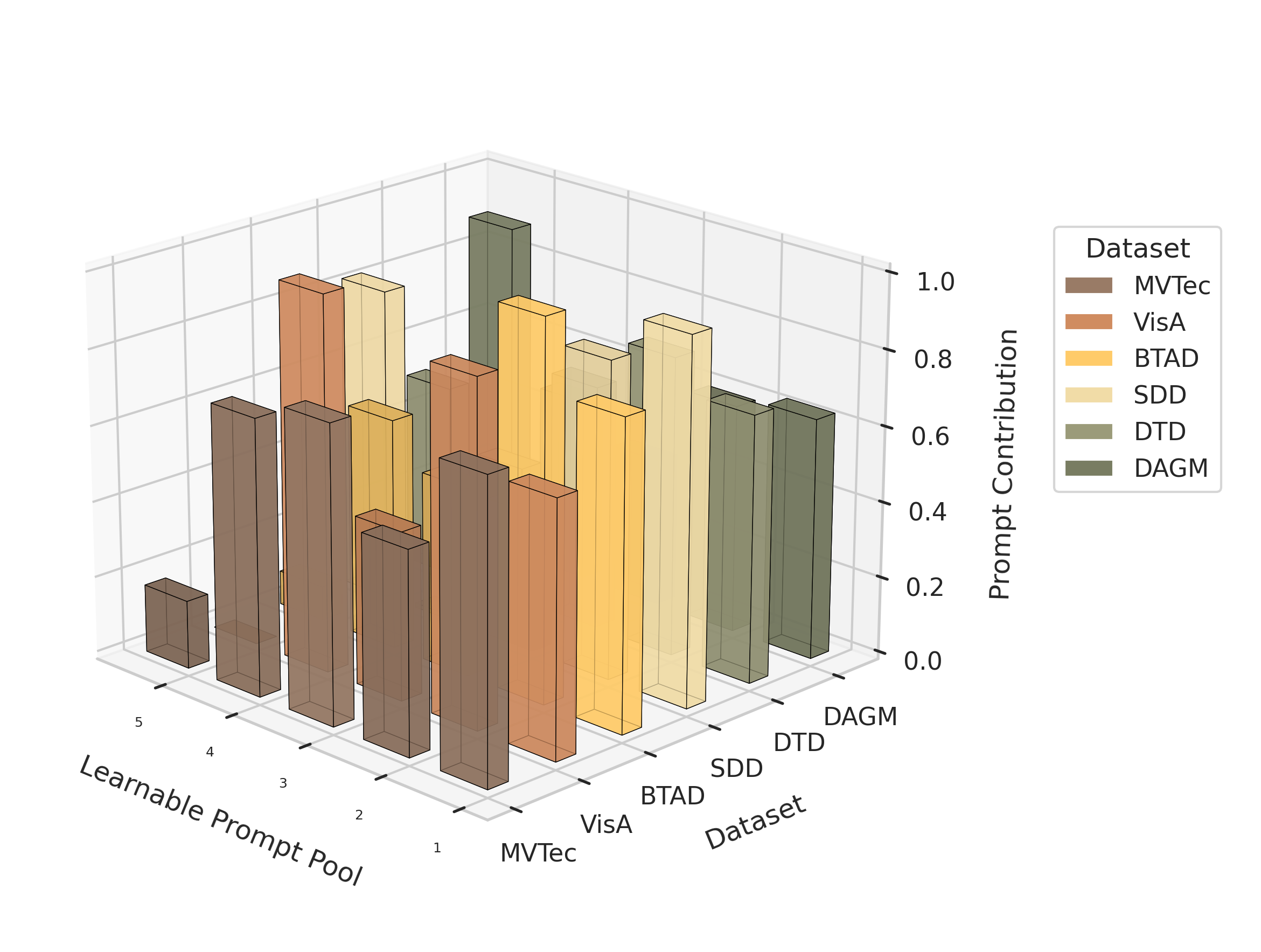}\label{fig:prompt_contribution}}
  \caption{
    (a) Prompt-pair similarity distributions: each horizontal boxplot shows absolute cosine similarity between learned prompt embeddings. Values close to 0 indicate high diversity. 
    (b) Prompt contribution: normalized contribution of a learnable prompt from the pool $\mathcal{P}$, aggregated over test images. Higher values indicate greater involvement during TTA.
  }
  \label{fig:prompt_analysis}
\end{figure}

To examine the diversity within our learnable prompt pool $\mathcal{P}$, we analyzed pairwise relationships among the learned prompt embeddings across multiple industrial datasets. Specifically, we measured the pairwise cosine similarity between prompt embeddings within each dataset. Figure~\ref{fig:prompt_diversity} summarizes these findings using boxplots. The results demonstrate consistent diversity across all evaluated datasets, with median similarity values generally below 0.5, indicating that the learned prompts capture distinct representations. \s{Prof Comment: NICE}

To further understand how this diversity is practically leveraged, Figure~\ref{fig:prompt_contribution} illustrates the relative contributions of each prompt in the pool across multiple datasets. The observed variations suggest that rather than relying on a single dominant prompt, the model adaptively emphasizes different prompts based on specific dataset characteristics. This adaptive utilization indicates that the prompt pool effectively leverages its inherent diversity to accommodate the varying requirements of different anomaly detection scenarios.

\begin{table}[h]
\centering
\caption{
Top-3 anomaly state matches for each learned prompt on the VisA dataset, retrieved via FAISS nearest neighbor search between the prompt embedding and the CLIP text embedding of ``a photo of a [anomaly state] object''. For each prompt, $\circ$ indicates matches after finetuning on the auxiliary set ($\mathcal{D}_{\mathrm{aux}}$), and $\bullet$ indicates matches after TTA on the target set ($\mathcal{D}_{\mathrm{tar}}$). The top-3 matches are listed in descending order of cosine similarity.
}
\label{tab:tta_prompt_matches_single}
\begin{tabular}{c|l}
\toprule
\textbf{Prompt} & \textbf{Top-3 Matches} \\
\midrule
1 & $\circ$ misaligned, bent, over-pressure \\
  & $\bullet$ bent, misaligned, flange misfit \\
\midrule
2 & $\circ$ abrasion mark, delamination, oxidized \\
  & $\bullet$ bent, flange misfit, clogged \\
\midrule
3 & $\circ$ damaged, corroded, weld defect \\
  & $\bullet$ damaged, impact dent, eroded \\
\midrule
4 & $\circ$ damaged, cracked, weld defect \\
  & $\bullet$ damaged, voltage spike damage, weld defect \\
\midrule
5 & $\circ$ damaged, paint peeling, weld defect \\
  & $\bullet$ damaged, voltage spike damage, contaminated \\
\bottomrule
\end{tabular}
\end{table}

Having established the inherent diversity within our learnable prompt pool $\mathcal{P}$, we further investigated how this diversity corresponds to meaningful anomaly characteristics expressed in natural language. To accomplish this, we conducted a nearest-neighbor analysis using FAISS~\cite{douze2024faiss}, comparing anomaly prompt embeddings ($T_a$) to embeddings generated from textual templates of the form “a photo of a [anomaly state] object.” Various anomaly state descriptors were used to represent potential real-world anomalies.

Table~\ref{tab:tta_prompt_matches_single} lists the top-3 nearest textual descriptors for each prompt embedding, both before and after TTA. Additional results are provided in Appendix~G. Prompts were initially trained on the MVTec auxiliary set and subsequently adapted to the VisA target set. The nearest-neighbor analysis revealed clear and meaningful alignments between the learned prompts and natural language anomaly descriptors. Prompts originally associated with anomaly states such as “misaligned,” “bent,” or “eroded” consistently maintained these semantic associations after adaptation, showing subtle shifts that reflect nuances in the target dataset. These observations underscore that the diversity within our learned prompt pool corresponds meaningfully to recognizable natural-language concepts.

\begin{table*}[h]
\centering
\caption{
Top-3 anomaly state matches for each learned prompt ($1$--$5$) on six datasets. 
Ranks are indicated by \ding{172}, \ding{173}, and \ding{174}. 
Each match is the closest to the template ``a photo of a [anomaly state] object''.
}
\label{tab:tta_prompt_matches_multi}
\footnotesize
\begin{tabularx}{\textwidth}{l|X|X|X|X|X|X}
\toprule
\textbf{Prompt} & \textbf{MVTec AD} & \textbf{VisA} & \textbf{BTAD} & \textbf{SDD} & \textbf{DAGM} & \textbf{DTD-Synthetic} \\
\midrule
1 & 
\ding{172} rusty \newline \ding{173} abrasion mark \newline \ding{174} flange misfit &
\ding{172} bent \newline \ding{173} misaligned \newline \ding{174} flange misfit &
\ding{172} overheated \newline \ding{173} short-circuited \newline \ding{174} misaligned &
\ding{172} bent \newline \ding{173} misaligned \newline \ding{174} electrical arcing &
\ding{172} bent \newline \ding{173} eroded \newline \ding{174} electrical arcing &
\ding{172} delamination \newline \ding{173} under-pressure \newline \ding{174} over-pressure \\
\midrule
2 & 
\ding{172} punctured \newline \ding{173} electrical arcing \newline \ding{174} short-circuited &
\ding{172} bent \newline \ding{173} flange misfit \newline \ding{174} clogged &
\ding{172} abrasion mark \newline \ding{173} weld defect \newline \ding{174} surface blistering &
\ding{172} abrasion mark \newline \ding{173} chipped \newline \ding{174} weld defect &
\ding{172} abrasion mark \newline \ding{173} grease smear \newline \ding{174} chipped &
\ding{172} electrical arcing \newline \ding{173} delamination \newline \ding{174} under-pressure \\
\midrule
3 & 
\ding{172} damaged \newline \ding{173} impact dent \newline \ding{174} cracked &
\ding{172} damaged \newline \ding{173} impact dent \newline \ding{174} eroded &
\ding{172} damaged \newline \ding{173} weld defect \newline \ding{174} abrasion mark &
\ding{172} damaged \newline \ding{173} abrasion mark \newline \ding{174} scratched &
\ding{172} surface fatigue crack \newline \ding{173} grease smear \newline \ding{174} scratched &
\ding{172} damaged \newline \ding{173} short-circuited \newline \ding{174} eroded \\
\midrule
4 & 
\ding{172} bent \newline \ding{173} gasket blown \newline \ding{174} cross-threaded &
\ding{172} damaged \newline \ding{173} voltage spike damage \newline \ding{174} weld defect &
\ding{172} damaged \newline \ding{173} cracked \newline \ding{174} broken &
\ding{172} weld defect \newline \ding{173} surface fatigue crack \newline \ding{174} damaged &
\ding{172} surface fatigue crack \newline \ding{173} cracked \newline \ding{174} damaged &
\ding{172} cracked \newline \ding{173} damaged \newline \ding{174} broken \\
\midrule
5 & 
\ding{172} scratched \newline \ding{173} grease smear \newline \ding{174} overheated &
\ding{172} damaged \newline \ding{173} voltage spike damage \newline \ding{174} contaminated &
\ding{172} damaged \newline \ding{173} weld defect \newline \ding{174} abrasion mark &
\ding{172} weld defect \newline \ding{173} damaged \newline \ding{174} surface fatigue crack &
\ding{172} surface fatigue crack \newline \ding{173} structural crack \newline \ding{174} weld defect &
\ding{172} surface fatigue crack \newline \ding{173} grease smear \newline \ding{174} structural crack \\
\bottomrule
\end{tabularx}
\end{table*}

Table~\ref{tab:tta_prompt_matches_multi} further extends this semantic exploration, reporting the top-3 nearest textual matches for each learned prompt ($1$--$5$) across all six industrial anomaly detection datasets. Each match is derived from the template "a photo of a [anomaly state] object."

This broader analysis underscores that individual prompts exhibit varying associations with different anomaly states depending on the dataset. Such variability demonstrates the model's capacity to adapt prompt representations across diverse domains. While some semantic associations remain stable, others shift considerably, reflecting both shared and domain-specific semantic cues. Collectively, these findings reinforce the effectiveness of our prompt pool in capturing diverse anomaly characteristics and adapting its semantic emphasis in response to domain-specific features.

\section{Medical anomaly localization results}
\label{app:Medical_localization}
\begin{table*}[h]
    \centering
    \caption{Comparison of model performance on medical anomaly localization datasets. 
    Pixel‑level results (AUROC, AUPRO) in parentheses. Best entries are marked with bold and second‑best entries with underline.}
    \begin{adjustbox}{width=\textwidth}
    \begin{tabular}{@{}lccccccc@{}}
        \toprule
        \multicolumn{8}{c}{\textbf{Medical Pixel-level Anomaly Localization (AUROC, AUPRO)}} \\
        \midrule
        Dataset        & SAA              & CoCoOp            & WinCLIP                   & APRIL-GAN                    & AnomalyCLIP                   & AdaCLIP                  & PILOT w/o TTA                 \\
        \midrule
        ISIC           & (83.8, 74.2)     & (51.7, 15.9)      & (83.3, 55.1)             & (\textbf{89.4}, \textbf{77.2}) & (87.4, \underline{74.5})     & (84.4, 54.5)             & (\underline{87.8}, 74.2)      \\
        CVC‑ColonDB    & (71.8, 31.5)     & (40.5,  2.6)      & (70.3, 32.5)             & (78.4, 64.6)                 & (\underline{88.5}, \textbf{79.3}) & (88.0, 63.9)             & (\textbf{88.6}, \underline{78.2}) \\
        CVC‑ClinicDB   & (66.2, 29.1)     & (34.8,  2.4)      & (51.2, 13.8)             & (80.5, 60.7)                 & (\underline{93.0}, \textbf{81.6}) & (\textbf{94.4}, 73.5)    & (92.1, \underline{78.5})      \\
        Kvasir         & (45.9, 13.3)     & (44.1,  3.5)      & (69.7, 24.5)             & (75.0, 36.2)                 & (93.2, \textbf{59.9})        & (\textbf{94.6}, 26.2)    & (\underline{94.0}, \underline{58.8}) \\
        \midrule
        Mean           & (66.9, 37.0)     & (42.8,  6.1)      & (68.6, 31.5)             & (80.8, 59.7)                 & (\underline{90.5}, \textbf{73.8}) & (90.4, 54.5)             & (\textbf{90.6}, \underline{72.4})    \\
        \bottomrule
    \end{tabular}
    \end{adjustbox}
    \label{table:medical_results}
\end{table*}

Medical anomaly localization results are reported in Table~\ref{table:medical_results}, with all evaluations conducted without TTA. Since medical datasets lack explicit “normal” examples, our TTA strategy cannot be applied; nevertheless, PILOT still produces competitive localization maps, achieving the highest mean AUROC and the second‑best AUPRO across benchmarks. These results underscore the generality of our dual‑branch prompt learning framework: even in the absence of adaptation, dynamically weighting prompts with the most relevant anomaly cues deliver state‑of‑the‑art performance in medical domains. 

\section{Computational Analysis}
\label{app:Computational_Analysis}

\s{Prof Comment: below, Gflops of Pilot are alculated in both train and TTA? --  sorry i think GFLOPs, latency, GPU Mem for all method are calculated during testing phase, only params are calculated for total model... do you think it would be better to report these values at all phases? maybe we can win there too.... -- Yes i can try this, but i think for industrial AD, people more like test time because it's a important application in real life. I see your point -- so i guess our train is not better than other methods? -- i think it's better than adaclip, but worse a little than anomalyclip, becuase AnomalyClIP is only single learnable prompt.. I see so just report test time but clarify in the caption. -- yes, i will}
\begin{table}[b]
  \centering
  \caption{Comparison of computational resource requirements for PILOT and main competitors. GFLOPs, latency, and GPU memory are reported for the test phase. Trainable parameters are reported for the training phase in all methods; for PILOT, test-time adaptation parameters are also shown (train / test).}
  \label{table:comp_analysis}
  \begin{adjustbox}{width=0.8\textwidth,center}
    \begin{tabular}{@{}lcccc@{}}
      \toprule
      \textbf{Method}
        & \textbf{Params (M)}
        & \textbf{GFLOPs}
        & \textbf{Latency (ms)}
        & \textbf{GPU Mem (MB)} \\
      \midrule
      AnomalyCLIP
        & 5.56
        & 482.06
        & 107.4
        & 2051.7 \\ 
      AdaCLIP
        & 10.67
        & 2872.37
        & 219.0
        & 8752.5 \\ 
      \midrule
      PILOT
        & 6.40 (Train) / 5.61 (Test)
        & 491.39
        & 109.9
        & 2398.4 \\ 
      \bottomrule
    \end{tabular}
  \end{adjustbox}
\end{table}

Table~\ref{table:comp_analysis} compares PILOT against our main competitors, AnomalyCLIP and AdaCLIP, in terms of model size, computational cost, inference speed, and GPU memory footprint. All measurements were obtained on an NVIDIA Ada A5000 GPU, as detailed in Appendix~\ref{app:PILOT_Details}. PILOT exhibits a resource profile closely aligned with AnomalyCLIP, and significantly outperforms AdaCLIP across all measured metrics. Importantly, PILOT employs a two-phase process: during the initial training phase, it utilizes a moderately sized model, while in the TTA phase, only the learnable prompt pool ($\mathcal{P}$) is updated. By limiting parameter updates to $\mathcal{P}$ during deployment, PILOT further reduces computational requirements, enhancing its practical suitability for near real-time anomaly detection in resource-constrained environments.

\section{Compared with AnovL}
\label{app:comp_AnovL}

AnovL~\cite{deng2023anovl} is, to our knowledge, the only existing test‑time adaptation method tailored for zero‑shot anomaly localization. It introduces a lightweight adapter on the visual token stream and refines pixel‑level anomaly maps by minimizing a self‑supervised reconstruction loss; the adapter learns to reconstruct features of normal patches, so that anomalous regions, whose reconstructions incur higher error, become more salient.

To investigate the complementarity of these adaptation schemes, we conducted an ablation in which PILOT’s image-level TTA is combined with AnovL’s pixel reconstruction loss (row “+ AnoVL Loss”). Although this hybrid approach retains PILOT’s adaptive prompt weighting mechanism and the visual adapter architecture, it failed to improve either localization or detection. This result suggests that the reconstruction and pseudo-label losses impose conflicting objectives on the same prompt parameters, generating contradictory update signals that destabilize adaptation and ultimately degrade performance.

By design, AnovL’s adaptation objective targets only pixel‑level refinement and does not update any image‑level scoring components. As a result, AnovL achieves modest gains in localization accuracy but has no impact on image‑level detection (Table~\ref{table:anovl}). In contrast, PILOT’s TTA strategy optimizes a high-confidence, image-level objective by adaptively updating the learnable prompt parameters according to their relevance for each test image, which leads to consistent improvements on both image- and pixel-level metrics. This dual benefit arises because PILOT emphasizes prompt parameters most aligned with the test data, while the fixed attribute embeddings act as semantic anchors to stabilize adaptation and prevent parameter drift.

To investigate the complementarity of these adaptation schemes, we conducted an ablation in which PILOT’s image‑level TTA is combined with AnovL’s pixel reconstruction loss (row “+ AnoVL Loss”). Although this hybrid approach retains PILOT’s adaptive prompt weighting mechanism and the visual adapter architecture, it failed to improve either localization or detection. We hypothesize that the reconstruction and pseudo‑label losses impose conflicting objectives on the same prompt parameters, driving contradictory update signals that destabilize adaptation and degrade performance.

\s{Prof Comment: Anovl is published where? do they report image level in their paper as well? you have cited its arxiv version.  -- anovl doesn't publish into conference as far as i know... but they have some citations.. and they do report image level performance without TTA-- weird!}

\begin{table*}[htbp]
  \centering
  \caption{Comparison between PILOT and AnovL on two benchmarks (MVTec AD and VisA). “w/o TTA” and “w/ TTA” indicate performance without and with test‑time adaptation, respectively. Arrows denote change under TTA: \Up (improvement), \Down (decline), \Same (no significant change, $\pm0.5$).}
  \label{table:anovl}
  \begin{adjustbox}{width=\textwidth,center}
    \begin{tabular}{@{}l 
        | cc cc 
        | cc cc@{}}
      \toprule
      \textbf{Method}
        & \multicolumn{4}{c|}{\textbf{MVTec AD}}
        & \multicolumn{4}{c}{\textbf{VisA}} \\
      \cmidrule(lr){2-5} \cmidrule(lr){6-9}
        & \multicolumn{2}{c}{Image (AUROC, AP)} 
        & \multicolumn{2}{c|}{Pixel (AUROC, AUPRO)} 
        & \multicolumn{2}{c}{Image (AUROC, AP)} 
        & \multicolumn{2}{c}{Pixel (AUROC, AUPRO)} \\
      \cmidrule(lr){2-3} \cmidrule(lr){4-5} \cmidrule(lr){6-7} \cmidrule(lr){8-9}
        & w/o TTA & w/ TTA 
        & w/o TTA & w/ TTA 
        & w/o TTA & w/ TTA 
        & w/o TTA & w/ TTA \\
      \midrule
      AnoVL
        & (87.9, 93.8)  & (87.9\Same, 93.8\Same)
        & (82.7, 60.2)  & (84.8\Up, 68.4\Up)
        & (72.6, 76.9)  & (72.6\Same, 76.9\Same)
        & (84.4, 62.0)  & (85.6\Up, 64.1\Up) \\
    \midrule

      \textbf{PILOT}
        & (90.2, 95.7)  & (92.1\Up, 96.6\Up)
        & (89.4, 79.2)  & (90.3\Up, 80.2\Up)
        & (82.0, 84.1)  & (84.1\Up, 85.7\Up)
        & (94.8, 85.6)  & (96.0\Up, 87.2\Up) \\

      + AnoVL Loss
        & (90.2, 95.7)  & (87.5\Down, 89.7\Down)
        & (89.4, 79.2)  & (66.4\Down, 52.3\Down)
        & (82.0, 84.1)  & (73.9\Down, 78.1\Down)
        & (94.8, 85.6)  & (79.4\Down, 65.8\Down) \\

      \bottomrule
    \end{tabular}
  \end{adjustbox}
\end{table*}
\section{Details of Attribute Memory Bank Construction}
\label{app:attribute_bank_construction}

To construct the attribute memory bank, we first define four fixed sets of text templates, each capturing domain-relevant semantics for anomaly detection. Specifically, two sets encode object states: a normal state set, which contains descriptors such as “normal \{\}” and “flawless \{\},” and an anomalous state set, which includes terms like “damaged \{\}” and “imperfect \{\}.” In addition, we curate three context template lists to capture diverse imaging scenarios: a set for general visual conditions (e.g., “a cropped photo of the \{\}” or “a blurry photo of a \{\} for anomaly detection”), an industrial context set (e.g., “an industrial photo of a \{\}” or “a bright industrial photo of the \{\}”), and a medical context set (e.g., “a CT scan of a \{\}” or “an ultrasound image of a \{\}”).

For each object class, we instantiate all possible combinations by inserting the class name into each entry of the normal and anomalous state sets, and further merging them with every template from the general, industrial, and medical context lists. This combinatorial process results in a comprehensive set of several hundred prompts per class, spanning a wide spectrum of visual conditions and attribute states. For example, for the class “bearing,” we synthesize prompts such as “normal bearing in a cropped photo,” “imperfect bearing in a bright industrial photo,” and “damaged bearing in an ultrasound image.”

All generated prompts are tokenized and encoded using the CLIP text encoder, after which the resulting embeddings are L2-normalized. To facilitate efficient retrieval, the embeddings are partitioned into two disjoint subsets corresponding to normal and anomalous states, thereby enabling the attribute memory bank to provide appropriate semantic vectors during prompt weighting for both image-level and pixel-level anomaly detection tasks. This construction ensures that the memory bank captures a rich and diverse set of attribute semantics, supporting robust prompt weighting and improved generalization to unseen domains.
\section{Pseudo Code}
\label{app:Pseudo_code}

\begin{algorithm}[htbp]
  \caption{Training Procedure for PILOT}
  \label{alg:pilot}
  \begin{algorithmic}[1]
    \REQUIRE Dataset $D_{aux} = \{(I_i, y_i, G_i)\}$; Learnable Prompt Pool $\mathcal{P}$; Attribute Memory Bank $\mathcal{U}$; Vision Encoder $f_{ visual}$; Text Encoder $f_{ text}$; epochs $E$; batch size $B$
    \STATE \textbf{Freeze} all weights except: last visual projection, learnable prompt pool, attribute memory bank \RComment{initialize model}
    \FOR{$e = 1$ \TO $E$}
      \FOR{each batch $(I, y, G)$ in $D_{aux}$}
        \STATE $x_{\rm cls}, \{X^{(m)}\}_{m \in M} \gets f_{ visual}(I)$ \RComment{extract global \& patch features from the vision encoder}
        \STATE $\alpha \gets Q_{lrn}(x_{\rm cls}, \mathcal{P})$
        \STATE $\bar p^n, \bar p^a \gets  \sum_{k=1}^K \alpha_k p_k^n,  \sum_{k=1}^K \alpha_k p_k^a$ \RComment{aggregate prompts with learned weights}
        \STATE $T_n, T_a \gets f_{ text}(\bar p^n, \bar p^a)$ \RComment{encode normal/anomaly prompt}
        \STATE 
        \STATE $\beta \gets Q_{ att}(x_{\rm cls}, \mathcal{U})$ 
        \STATE initial Gumbel noise
        \STATE $\phi_{c} \gets \operatorname{entmax}_{1.5}\bigl(\beta_{c} + \text{noise}\bigr)$
        \RComment{ensure sparsity}
        \STATE $T_{att} \gets \sum_{c=1}^{C} \phi_{c}\,u_{c}$         \RComment{construct attribute representation}
        \STATE 
        \STATE $T_n^{\rm fus}, T_a^{ fus} \gets \mathrm{Projector}(T_{ att}, T_n, T_a)$ \RComment{fuse by projector}
        \STATE $\hat{y} = \sigma(sim(x_{\rm cls}, T_{(\cdot)}^{ fus}))$ \RComment{image-level score}
        \STATE $\mathcal{L}_{\rm image} = \mathrm{BCE}(\hat{y}, y)$ \RComment{image-level loss}
        \FOR{each $m \in M$}
          \STATE $\hat{G}^{(m)} = \sigma(sim(X^{(m)}, T_{(\cdot)}^{\rm fus}))$ \RComment{pixel anomaly map}
        \ENDFOR
        \STATE $\mathcal{L}_{\rm pixel} = \frac{1}{|M|} \sum_{m \in M} [\mathrm{Focal}(\hat{G}^{(m)}, G) + \mathrm{Dice}(\hat{G}^{(m)}, G)]$ \RComment{pixel-level loss}
        \STATE $\mathcal{L}_{\rm PILOT} = \mathcal{L}_{\rm image} + \mathcal{L}_{\rm pixel}$ \RComment{total loss}
        \STATE \textbf{Update} prompt pool, attribute bank, projection using $\mathcal{L}_{\rm PILOT}$
      \ENDFOR
    \ENDFOR
  \end{algorithmic}
\end{algorithm}

In Appendix~\ref{app:Pseudo_code}, we present high‑level pseudocode for both the training and test‑time adaptation procedures of PILOT. Algorithm~\ref{alg:pilot} details the end‑to‑end optimization on the auxiliary dataset \(D_{\mathrm{aux}}\), where we jointly update the learnable prompt pool \(\mathcal{P}\), attribute memory bank \(\mathcal{U}\), and final visual projection layer using combined image‑level and pixel‑level objectives. Algorithm~\ref{alg:pilot_tta} then describes our confidence‑based, image-level objective TTA scheme on an unlabeled target set \(D_{\mathrm{tar}}\), in which only the relevant prompt parameters and their associated $r_k$, $s_k$ embeddings are fine‑tuned to align with high‑confidence pseudo‑labels.

\begin{algorithm}[htbp]
  \caption{Test-Time Adaptation Procedure for PILOT}
  \label{alg:pilot_tta}
  \begin{algorithmic}[1]
    \REQUIRE Target dataset $D_{tar} = \{I_t\}$; Trained model ($f_{visual}$, $f_{text}$, prompt pool $\mathcal{P}$, attribute bank $\mathcal{U}$); confidence fraction $\rho$; epochs $E$
    \STATE \textbf{Freeze} all weights except selected prompt(s) and associated selector/prototype embeddings
    \FOR{$e = 1$ \TO $E$}
      \FOR{each $I_t$ in $D_{tar}$}
        \STATE $x_{\rm cls}^t \gets f_{visual}(I)$ \RComment{extract global features}
        \STATE $\bar p^n, \bar p^a \gets  \sum_{k=1}^K \alpha_k p_k^n,  \sum_{k=1}^K \alpha_k p_k^a$ \RComment{aggregate prompts with learned weights}
        \STATE $T_n, T_a \gets f_{ text}(\bar p^n, \bar p^a)$ \RComment{encode normal/anomaly prompt}
        \STATE $\beta \gets Q_{ att}(x_{\rm cls}, \mathcal{U})$ 
        \STATE initial Gumbel noise
        \STATE $\phi_{c} \gets \operatorname{entmax}_{1.5}\bigl(\beta_{c} + \text{noise}\bigr)$
        \RComment{ensure sparsity}
        \STATE $T_{ att} \gets \sum_{c=1}^{C} \phi_{c}\,u_{c}$         \RComment{construct attribute representation}
        \STATE $T_n^{\rm fus}, T_a^{ fus} \gets \mathrm{Projector}(T_{ att}, T_n, T_a)$ \RComment{fuse by projector}
        \STATE $\tilde{y} = \sigma(\mathrm{sim}(x_{\rm cls}^t, T_{(\cdot)}^{fus}))$ \RComment{image-level anomaly score}
      \ENDFOR
      \STATE Collect all scores $\{\tilde{y}_i\}_{i=1}^{|D_{tar}|}$
      \STATE Identify top/bottom $\rho$-fraction: $A = \mathrm{Top}_k\{\tilde{y}_i\}$, $N = \mathrm{Bot}_k\{\tilde{y}_i\}$, $k = \lceil \rho |D_{tar}| \rceil$
      \STATE Assign pseudo-labels: $\tilde{y}_i=1$ if $i\in A$, $\tilde{y}_i=0$ if $i\in N$
      \STATE $\mathcal{L}_{image}^{TTA} = \frac{1}{|A|+|N|} \sum_{i \in A \cup N} \mathrm{BCE}(\tilde{y}_i, \text{pseudo-labels})$ \RComment{image-level TTA loss}
      \STATE \textbf{Update} only the prompt(s) and associated embedding relevant to the current input to minimize $\mathcal{L}_{\text{image}}^{\text{TTA}}$
    \ENDFOR
  \end{algorithmic}
\end{algorithm}
\section{Visualization}
\label{app:visualization}
\begin{figure*}[h]
  \centering
  \includegraphics[width=0.8\linewidth]
  {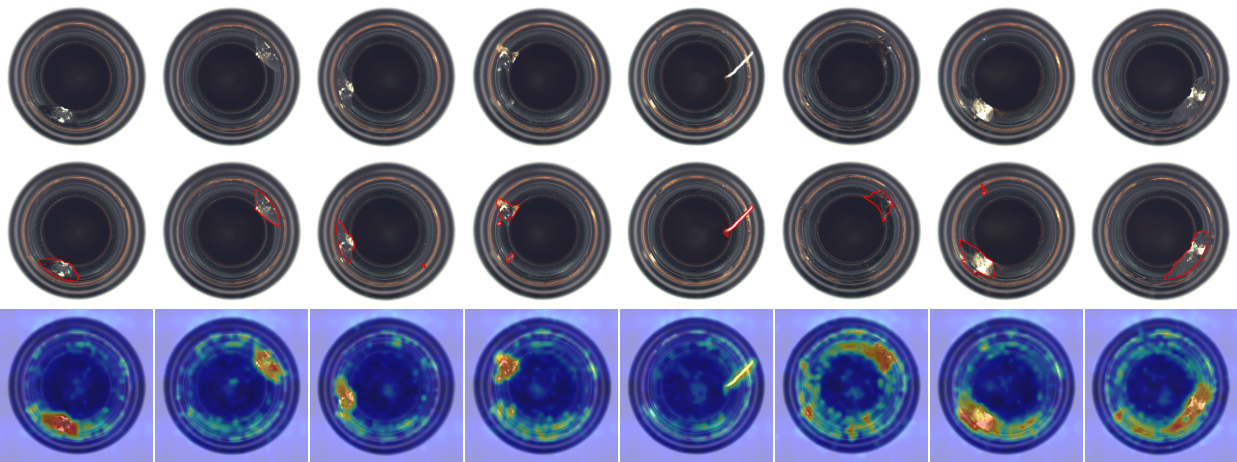}
  \caption{Visualization of input images (top), ground‐truth masks (middle), and model‐generated anomaly maps (bottom) for the \textit{bottle} category in the MVTec AD dataset. All results are generated by PILOT.}
  \label{fig:mvtec_bottle}
\end{figure*}

\begin{figure*}[h]
  \centering
  \includegraphics[width=0.8\linewidth]{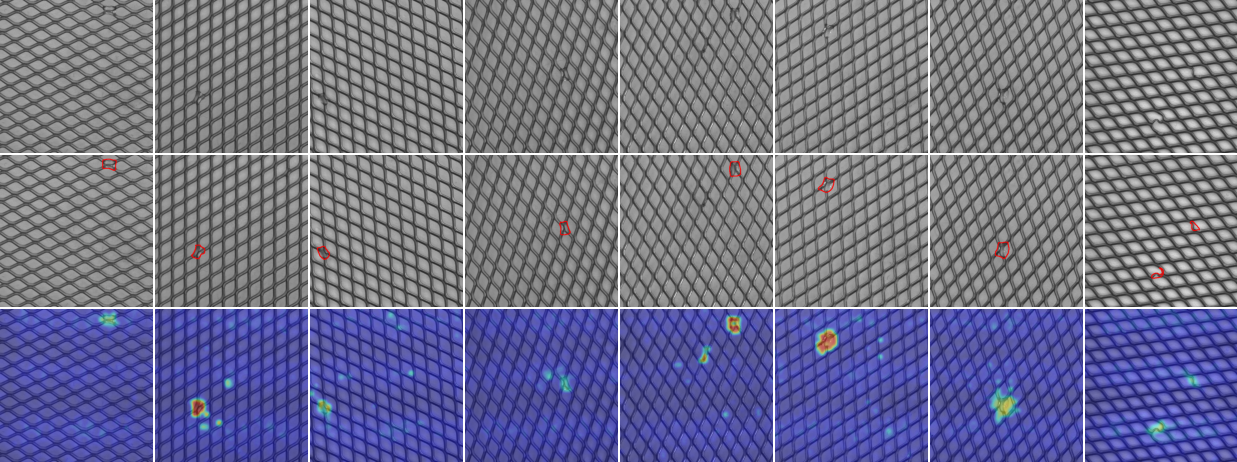}
  \caption{Visualization of input images (top), ground‐truth masks (middle), and model‐generated anomaly maps (bottom) for the \textit{grid} category in the MVTec AD dataset. All results are generated by PILOT.}
  \label{fig:mvtec_grid}
\end{figure*}

\begin{figure*}[h]
  \centering
  \includegraphics[width=0.8\linewidth]{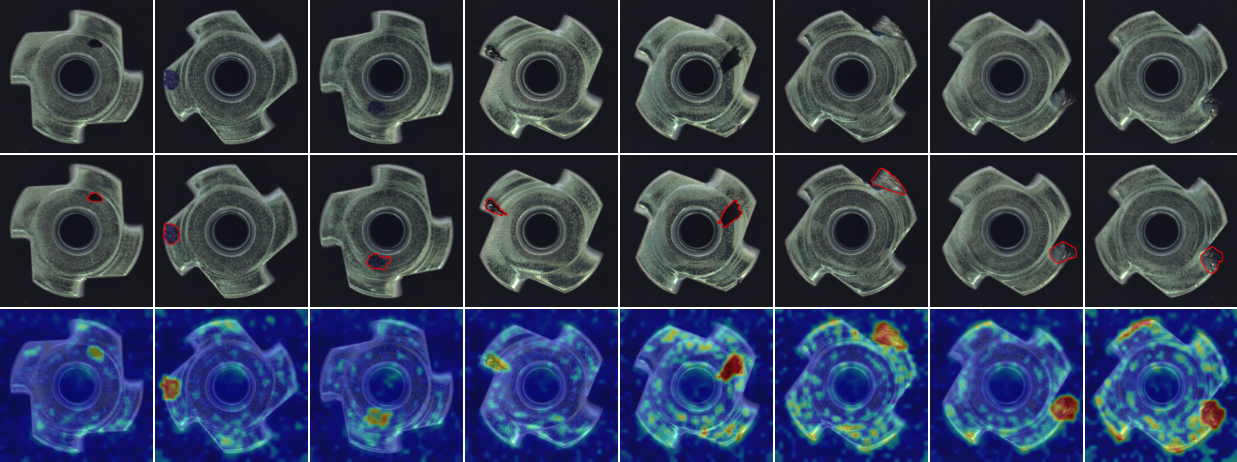}
  \caption{Visualization of input images (top), ground‐truth masks (middle), and model‐generated anomaly maps (bottom) for the \textit{metal nut} category in the MVTec AD dataset. All results are generated by PILOT.}
  \label{fig:mvtec_nut}
\end{figure*}

\begin{figure*}[h]
  \centering
  \includegraphics[width=0.8\linewidth]{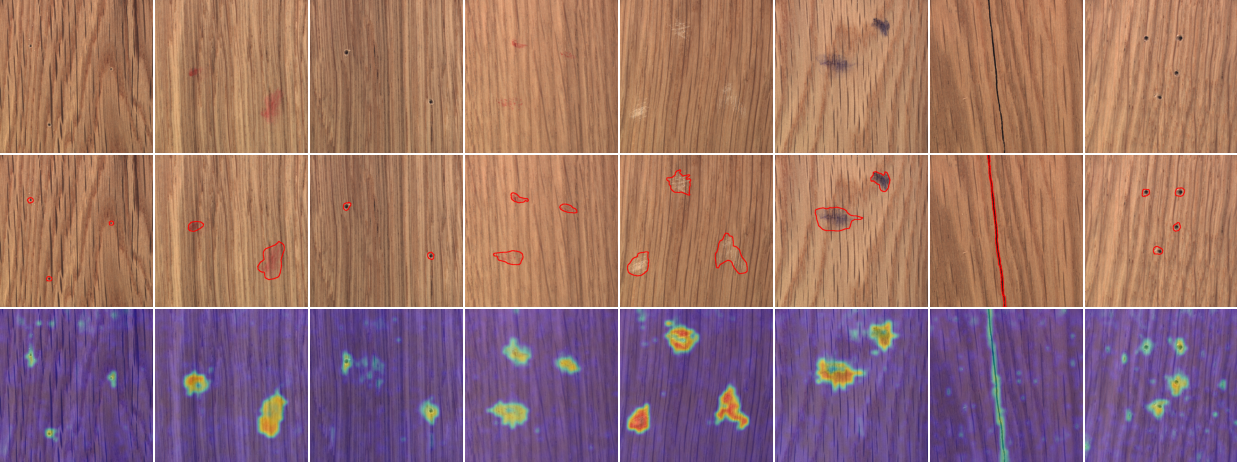}
  \caption{Visualization of input images (top), ground‐truth masks (middle), and model‐generated anomaly maps (bottom) for the \textit{wood} category in the MVTec AD dataset. All results are generated by PILOT.}
  \label{fig:mvtec_wood}
\end{figure*}

\begin{figure*}[h]
  \centering
  \includegraphics[width=0.8\linewidth]{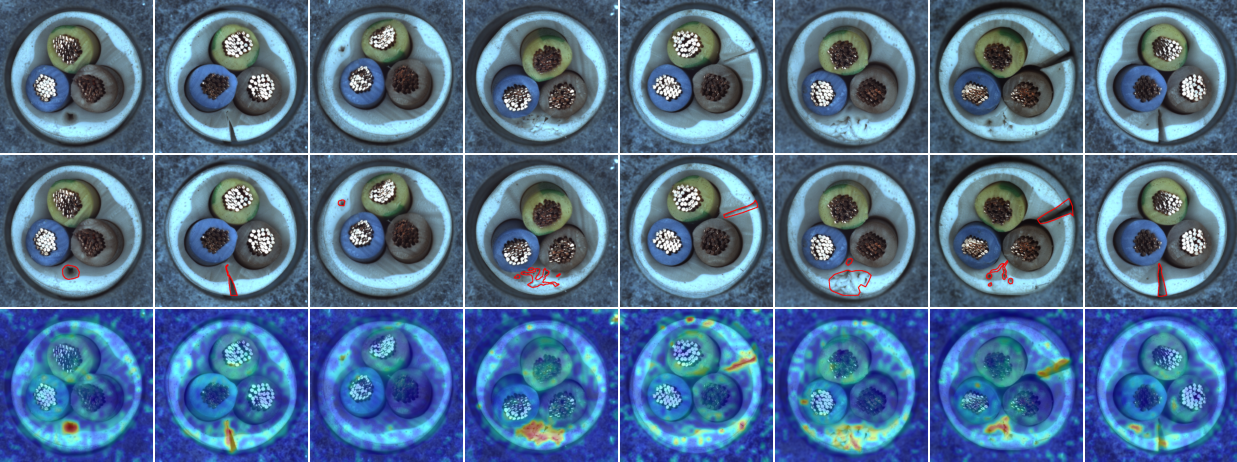}
  \caption{Visualization of input images (top), ground‐truth masks (middle), and model‐generated anomaly maps (bottom) for the \textit{cable} category in the MVTec AD dataset. All results are generated by PILOT.}
  \label{fig:mvtec_cable}
\end{figure*}

In this section, we present qualitative examples for a subset of categories from each industrial dataset generated by PILOT. Figures~\ref{fig:mvtec_bottle}-\ref{fig:dtd_Perforated} illustrate representative images and corresponding anomaly maps; to maintain clarity, we omit exhaustive visualizations of every class.

\s{in all below Figs, in the caption, mention that these are the results of Pilot. -- OKay    WHat makes all these more interesting, if you could show the decoded prompts -- as some sort of text explanation when anomaly is detected. You perhaps can provide these for these figures? and also would be nice to include normal samples as well to show that you don't detect anomalies in normal samples and can provide a nice informative explanation. what do you think? -- yes, i like this idea, but i am still thinking how to do it  this will be an addition to the novel parts of the paper that makes it attractive from explainability point of view .. -- yes, i will try my best to see if this experiment can be done, currently i am thinking that compared encode learnable prompt with ftext, and compared with other template like "A photo of a {anomaly} etc", and to see if the embedding is close and maybe they have relationship between each other and that learnable prompt can be consider as this kind of anomaly?   have you also thought about just simply give the Tnfus a and Tafus to a decoder and see what they generate?  and perhaps Tn/Ta and Tatt-- The problem is clip doesn't have decoder.. and because the training data for clip is not public so i was wondering is there any public decoder available online?     I was thinking of having a decoder trained on the datset we have in the fine tuning phase (perhaps a simple decdoer with limited outputs aligned with our small dictionary we have ... but for sure having a pretarined one then fine tuning it would be much more efficient.... you may use other clip-based decoders... not the exact one... regardless of how we generate these text, it will be super interesting and eye catching and i suggest even showing it on the first page of the paper. I believe none of ZSAD methods have done it, right? -- yes, i can definitely try it, but it would take a lot of time to do as it's sounds complex... i want to firstly finish the revision of the paper, then try this . Okay, first complete the suggested revisions till tomorrow, then see if you can    from  chatGPT: Try fine-tuning a small autoregressive language model (like GPT-2 or T5) on CLIP text embeddings + their corresponding text. It’s relatively lightweight and gives reasonable results. Or, use the nearest-neighbor retrieval approach for fast prototyping and debugging.   OR any other techniques. This makes a nice explanation on the method as it is natural to have such an explanation, but everyone has avoided this due to their wherever! -- yes, i agree it would be a good explanation for the paper}

\s{
\r{Here is what chatgpt suggest. the first option is similar to what we have, note that we have a limited dictionary}
\t{You're right to notice that while CLIP (Contrastive Language--Image Pretraining) provides strong image-to-text and text-to-image embedding alignment, it does not include a native ``decoder'' that can turn CLIP text embeddings back into natural language. This is a known limitation of the architecture---it's built for similarity comparison rather than generative decoding.}
That said, here are some potential directions and methods you can explore to ``decode'' or generate natural language from CLIP text embeddings:
\begin{enumerate}[leftmargin=*]
    \item \textbf{Training a Text Decoder on CLIP Text Embeddings} \\
    If you already have access to CLIP text embeddings (from the text encoder), you could:
    \begin{itemize}
        \item Train a lightweight language decoder (e.g., GPT-2, T5) to map CLIP embeddings to text sequences.
        \item You'll need a dataset of (CLIP\_embedding, original\_text) pairs to fine-tune the decoder.
    \end{itemize}
    This approach learns to approximate the inverse of the CLIP text encoder, but it won't be perfect because the CLIP embedding space is not bijective---some text details are lost in compression.
    \item \textbf{Use CLIP with a Captioning Model (Indirect Method)} \\
    If you're working with image CLIP embeddings, you can:
    \begin{itemize}
        \item Use the CLIP image encoder embedding as input to a decoder from an image captioning model like BLIP, Flamingo, or even a Vision-to-Text T5 variant.
        \item Some models (e.g., BLIP-2, GIT) can take CLIP-style visual embeddings and output text.
    \end{itemize}
    But this only works if your input is a CLIP image embedding, not a text embedding.
    \item \textbf{Leverage Nearest Neighbor in a Vector Database} \\
    A non-generative, yet simple and useful method:
    \begin{itemize}
        \item Store a large number of known text embeddings (paired with their texts) in a vector database like FAISS.
        \item For a given CLIP embedding, retrieve the nearest matching text (or top-N).
    \end{itemize}
    Pros: No need for training. \\
    Cons: Not generative, and depends on corpus size and coverage.
    \item \textbf{Use Multimodal Diffusion or Inversion Methods (Experimental)} \\
    Recent research in embedding inversion might help, e.g.:
    \begin{itemize}
        \item ``UnCLIP'' from OpenAI's DALL$\cdot$E 2 architecture: it uses a CLIP text embedding $\rightarrow$ image $\rightarrow$ caption pipeline.
        \item ``CLIPText2Text'' or similar: some works try to invert CLIP embeddings via diffusion models or learned priors.
    \end{itemize}
    These are research-heavy and usually not plug-and-play.
\end{enumerate}
\textbf{Recommended Strategy for You} \\
If you want something practical:
\begin{itemize}
    \item Try fine-tuning a small autoregressive language model (like GPT-2 or T5) on CLIP text embeddings plus their corresponding text. It's relatively lightweight and gives reasonable results.
    \item Or, use the nearest-neighbor retrieval approach for fast prototyping and debugging.
\end{itemize}
Let me know your dataset or task setting (e.g., are these embeddings from text prompts? image captions? prompts for generation?) and I can suggest a more targeted pipeline.}


\begin{figure*}[h]
  \centering
  \includegraphics[width=0.8\linewidth]{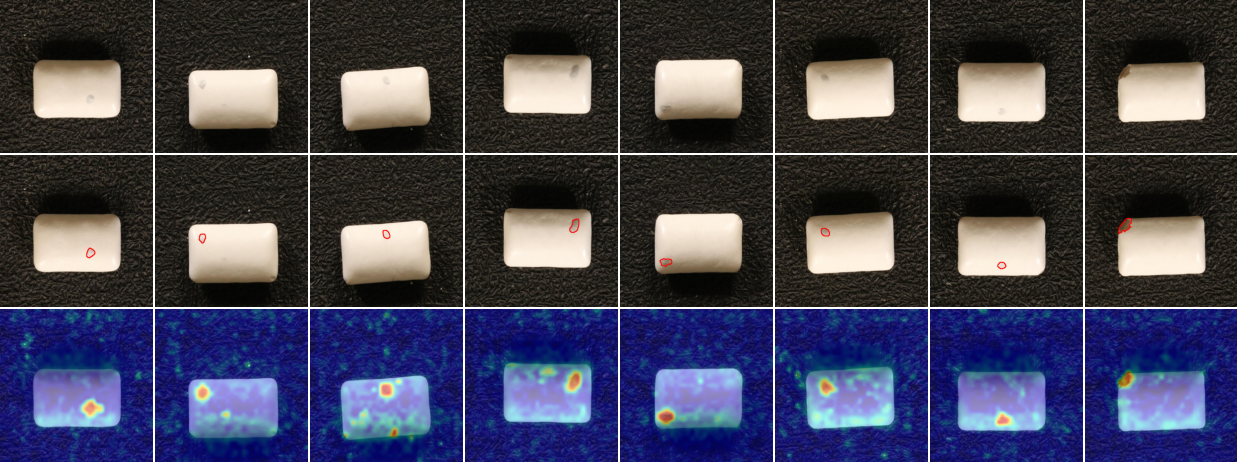}
  \caption{Visualization of input images (top), ground‐truth masks (middle), and model‐generated anomaly maps (bottom) for the \textit{chewinggum} category in the VisA dataset. All results are generated by PILOT.}
  \label{fig:}
\end{figure*}
\begin{figure*}[h]
  \centering
  \includegraphics[width=0.8\linewidth]{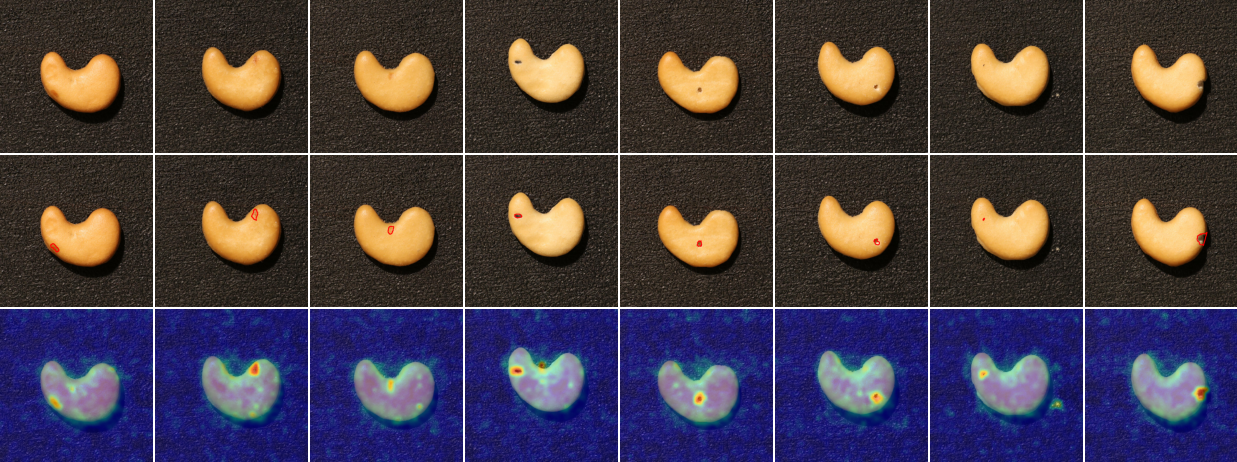}
  \caption{Visualization of input images (top), ground‐truth masks (middle), and model‐generated anomaly maps (bottom) for the \textit{cashew} category in the VisA dataset. All results are generated by PILOT.}
  \label{fig:}
\end{figure*}
\begin{figure*}[h]
  \centering
  \includegraphics[width=0.8\linewidth]{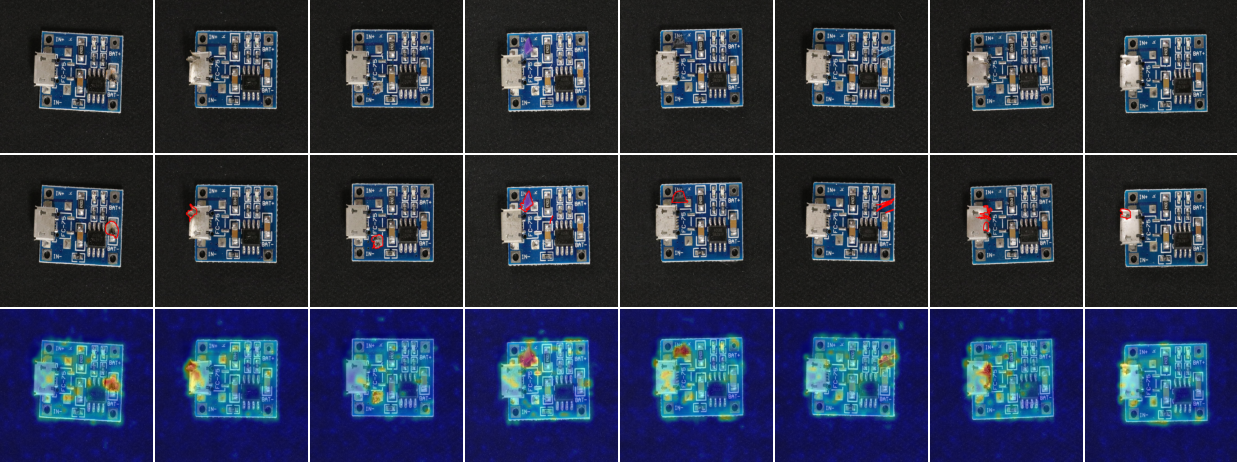}
  \caption{Visualization of input images (top), ground‐truth masks (middle), and model‐generated anomaly maps (bottom) for the \textit{pcb4} category in the VisA dataset. All results are generated by PILOT.}
  \label{fig:}
\end{figure*}
\begin{figure*}[h]
  \centering
  \includegraphics[width=0.8\linewidth]{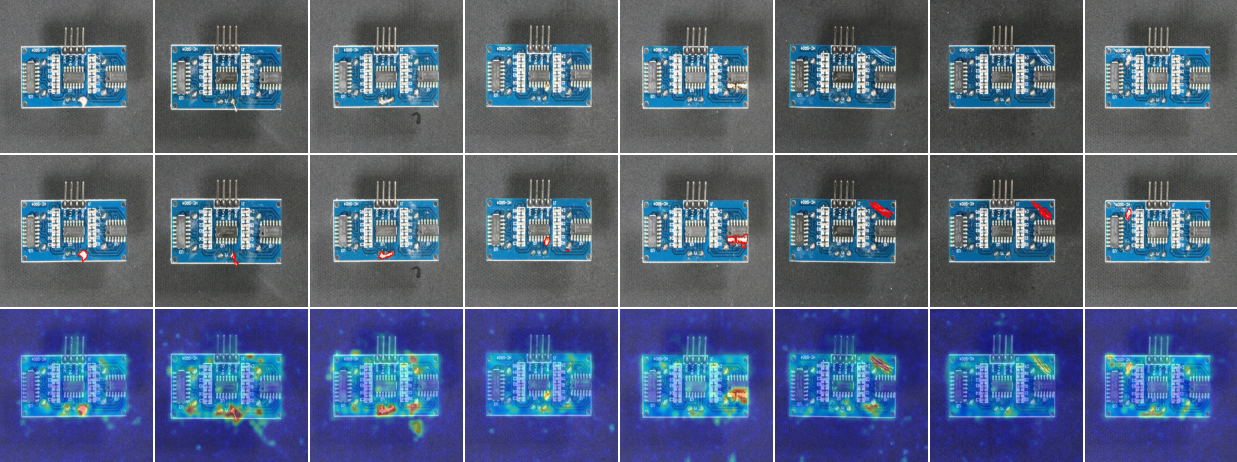}
  \caption{Visualization of input images (top), ground‐truth masks (middle), and model‐generated anomaly maps (bottom) for the \textit{pcb2} category in the VisA dataset. All results are generated by PILOT.}
  \label{fig:}
\end{figure*}
\begin{figure*}[h]
  \centering
  \includegraphics[width=0.8\linewidth]{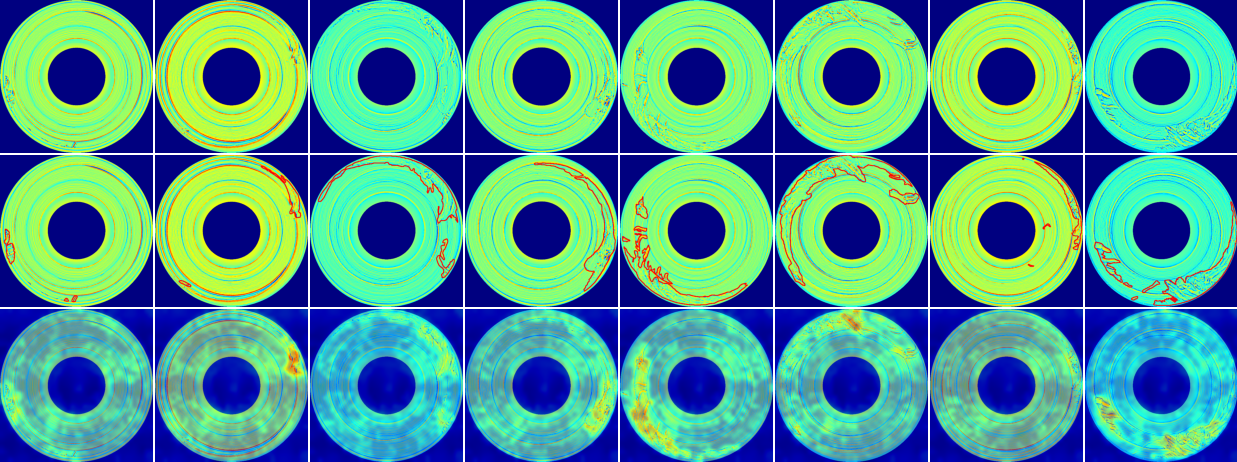}
  \caption{Visualization of input images (top), ground‐truth masks (middle), and model‐generated anomaly maps (bottom) for the \textit{01} category in the BTAD dataset. All results are generated by PILOT.}
  \label{fig:}
\end{figure*}
\begin{figure*}[h]
  \centering
  \includegraphics[width=0.8\linewidth]{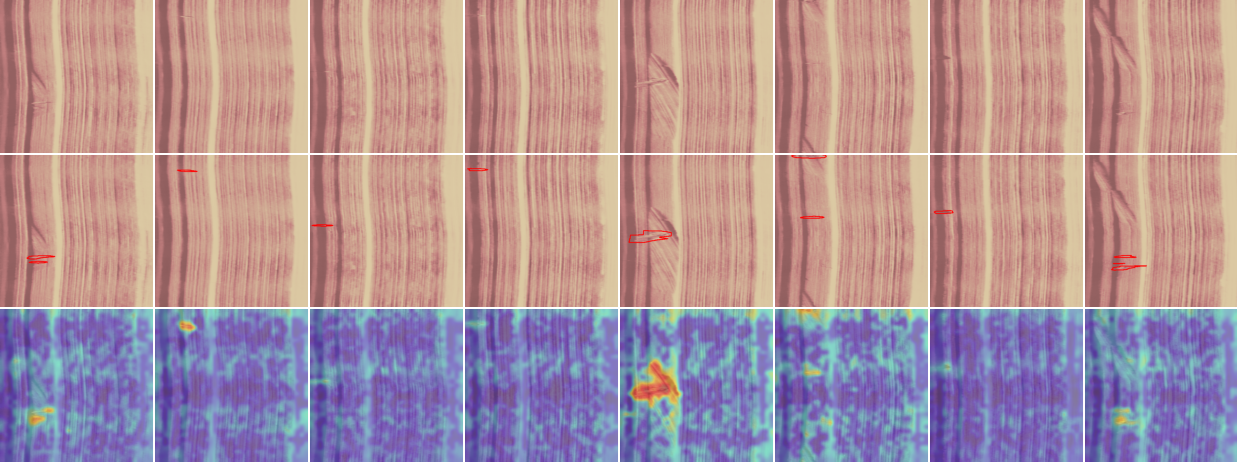}
  \caption{Visualization of input images (top), ground‐truth masks (middle), and model‐generated anomaly maps (bottom) for the \textit{02} category in the BTAD dataset. All results are generated by PILOT.}
  \label{fig:}
\end{figure*}
\begin{figure*}[h]
  \centering
  \includegraphics[width=0.8\linewidth]{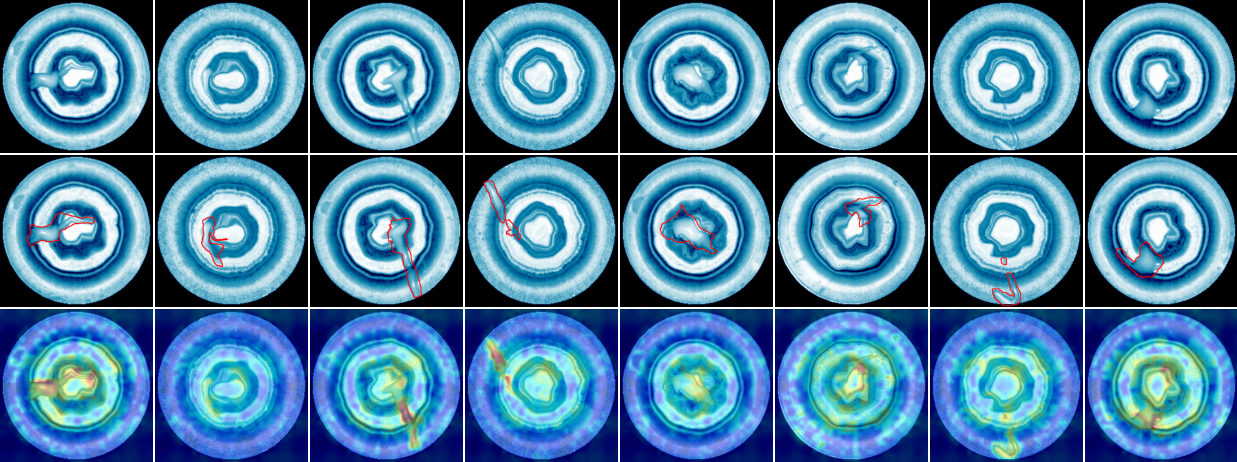}
  \caption{Visualization of input images (top), ground‐truth masks (middle), and model‐generated anomaly maps (bottom) for the \textit{03} category in the BTAD dataset. All results are generated by PILOT.}
  \label{fig:}
\end{figure*}
\begin{figure*}[h]
  \centering
  \includegraphics[width=0.8\linewidth]{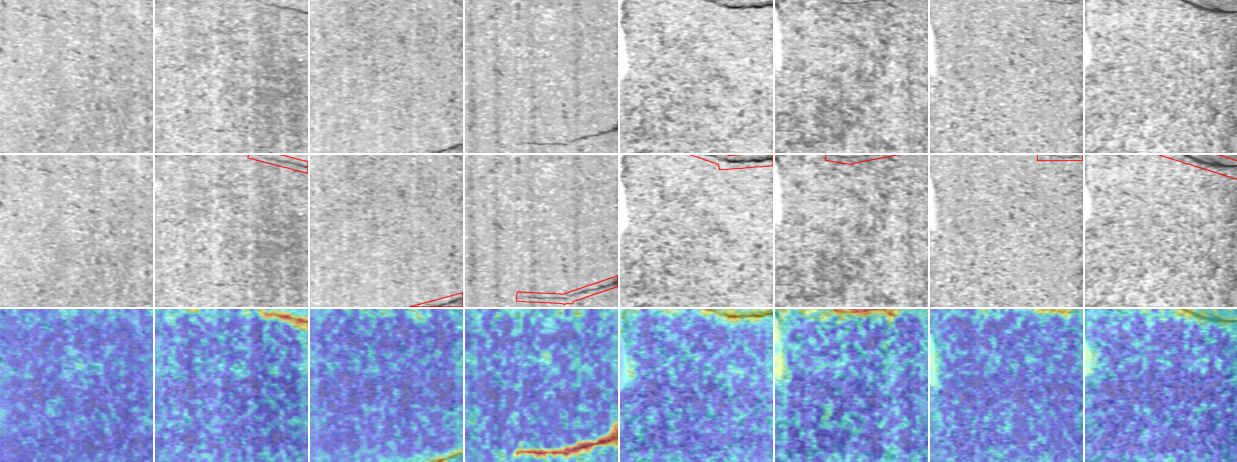}
  \caption{Visualization of input images (top), ground‐truth masks (middle), and model‐generated anomaly maps (bottom) for the \textit{electrical commutators} category in the SDD dataset. All results are generated by PILOT.}
  \label{fig:}
\end{figure*}
\begin{figure*}[h]
  \centering
  \includegraphics[width=0.8\linewidth]{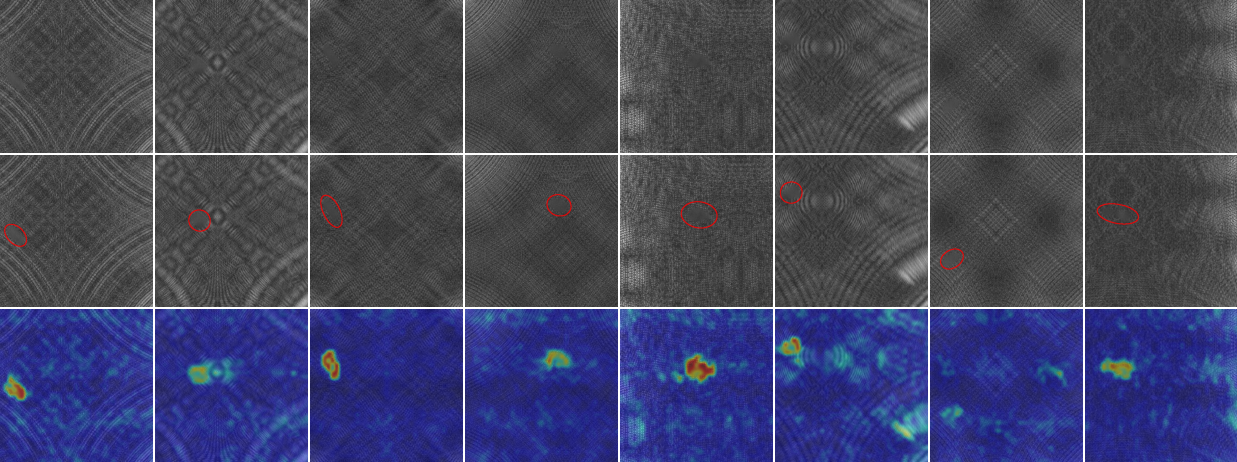}
  \caption{Visualization of input images (top), ground‐truth masks (middle), and model‐generated anomaly maps (bottom) for the \textit{class 1} category in the DAGM dataset. All results are generated by PILOT.}
  \label{fig:}
\end{figure*}
\begin{figure*}[h]
  \centering
  \includegraphics[width=0.8\linewidth]{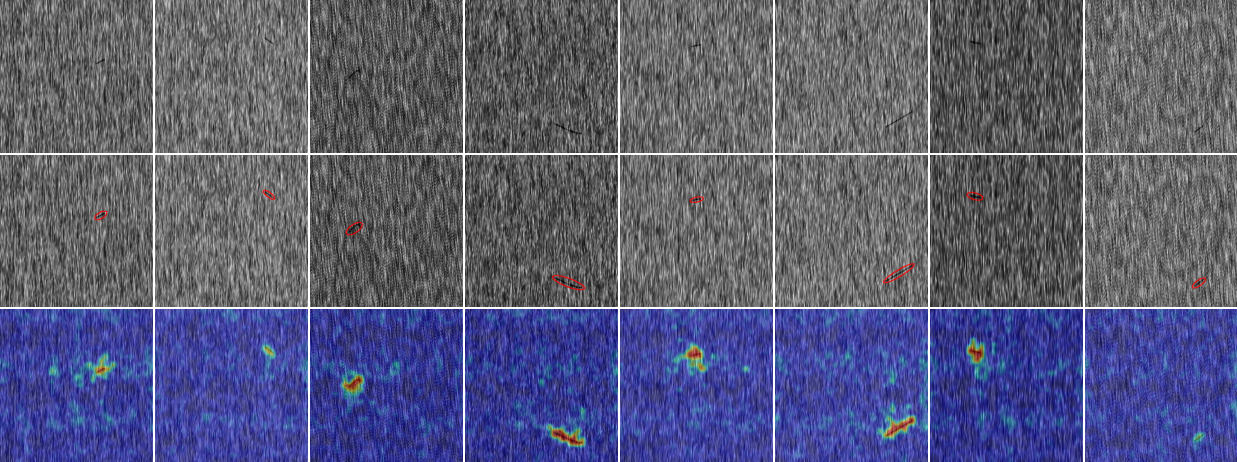}
  \caption{Visualization of input images (top), ground‐truth masks (middle), and model‐generated anomaly maps (bottom) for the \textit{class 2} category in the DAGM dataset. All results are generated by PILOT.}
  \label{fig:}
\end{figure*}
\begin{figure*}[h]
  \centering
  \includegraphics[width=0.8\linewidth]{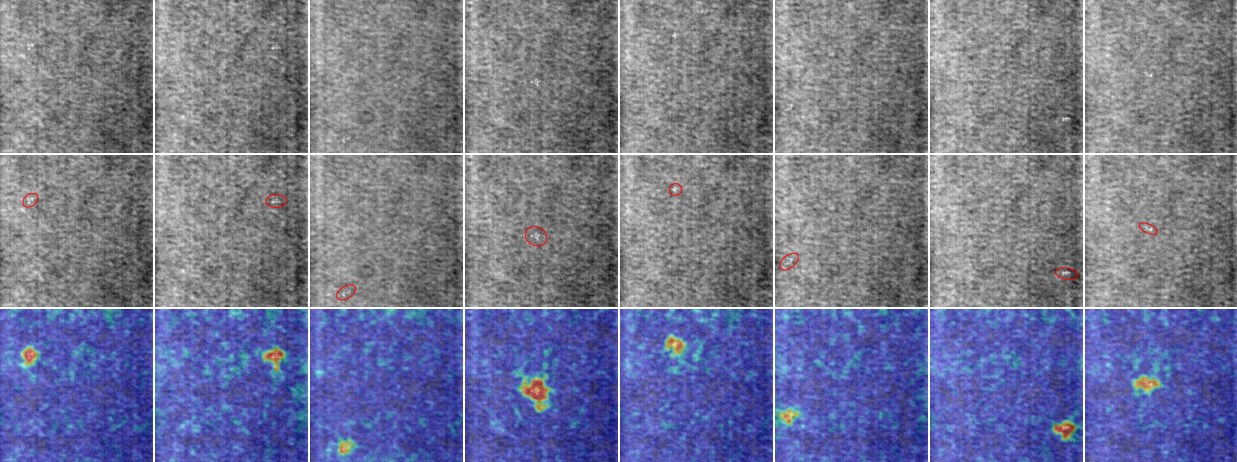}
  \caption{Visualization of input images (top), ground‐truth masks (middle), and model‐generated anomaly maps (bottom) for the \textit{class 3} category in the DAGM dataset. All results are generated by PILOT.}
  \label{fig:}
\end{figure*}
\begin{figure*}[h]
  \centering
  \includegraphics[width=0.8\linewidth]{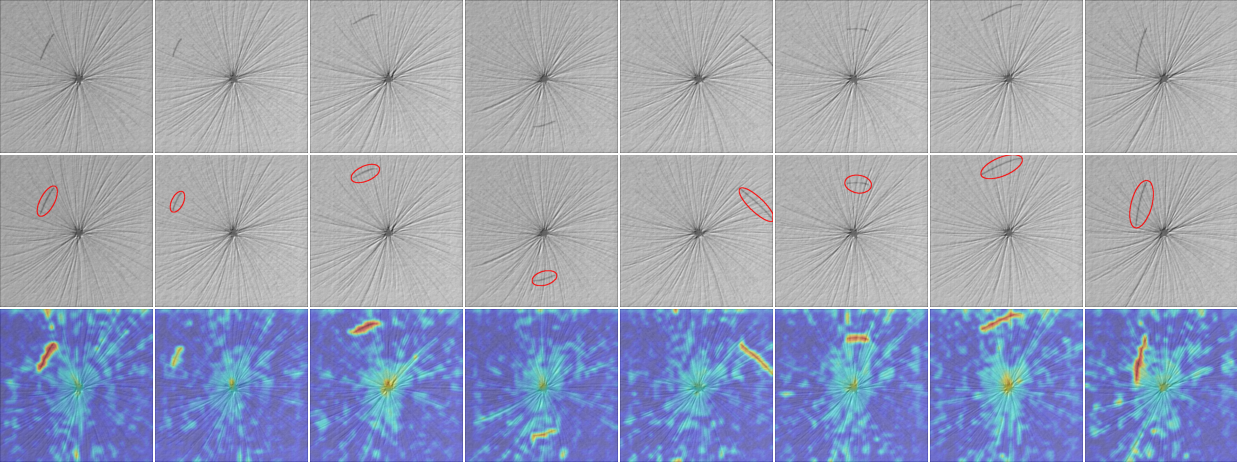}
  \caption{Visualization of input images (top), ground‐truth masks (middle), and model‐generated anomaly maps (bottom) for the \textit{class 4} category in the DAGM dataset. All results are generated by PILOT.}
  \label{fig:}
\end{figure*}
\begin{figure*}[h]
  \centering
  \includegraphics[width=0.8\linewidth]{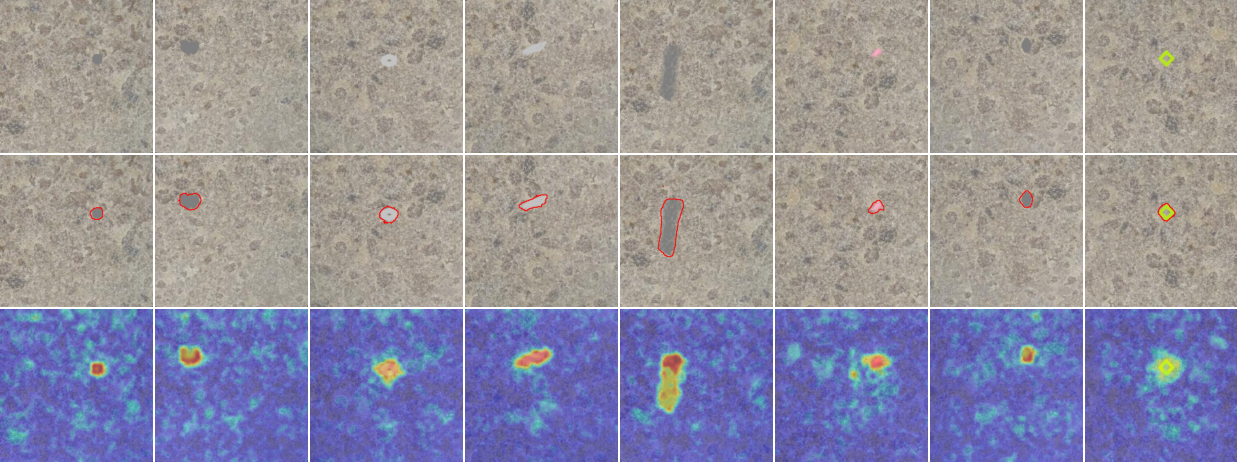}
  \caption{Visualization of input images (top), ground‐truth masks (middle), and model‐generated anomaly maps (bottom) for the \textit{Blotchy} category in the DTD dataset. All results are generated by PILOT.}
  \label{fig:}
\end{figure*}
\begin{figure*}[h]
  \centering
  \includegraphics[width=0.8\linewidth]{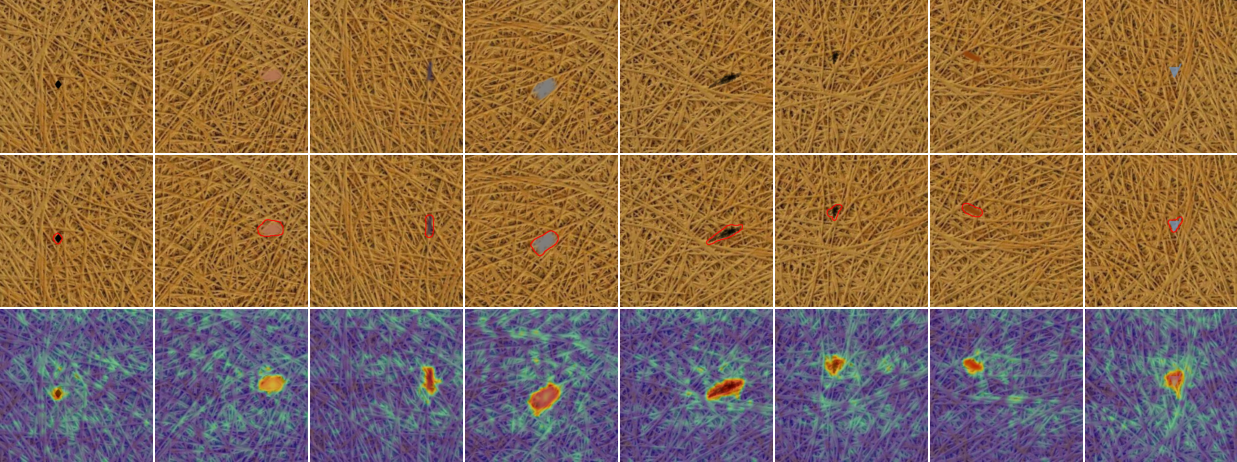}
  \caption{Visualization of input images (top), ground‐truth masks (middle), and model‐generated anomaly maps (bottom) for the \textit{Fibrous} category in the DTD dataset. All results are generated by PILOT.}
  \label{fig:}
\end{figure*}
\begin{figure*}[h]
  \centering
  \includegraphics[width=0.8\linewidth]{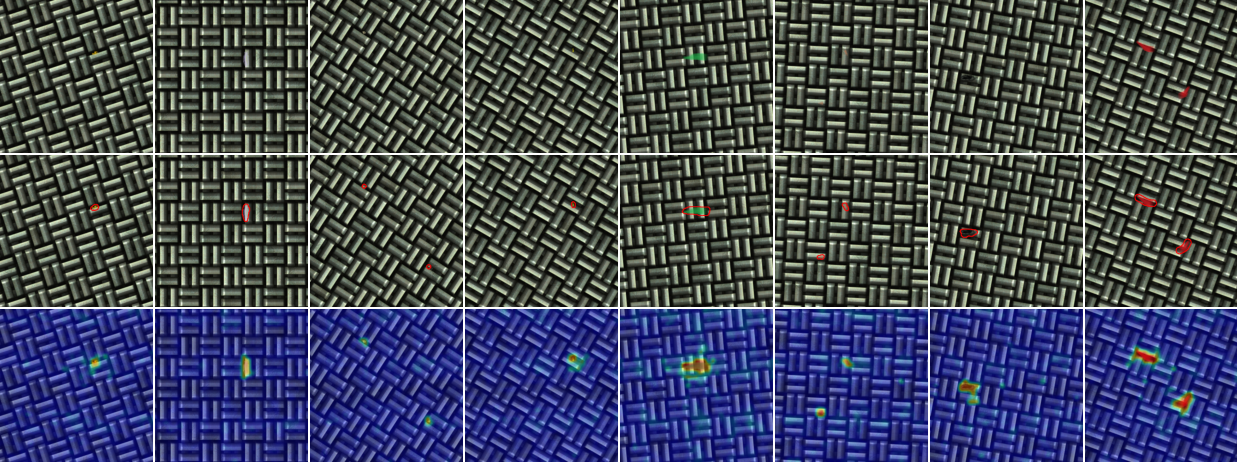}
  \caption{Visualization of input images (top), ground‐truth masks (middle), and model‐generated anomaly maps (bottom) for the \textit{Woven} category in the DTD dataset. All results are generated by PILOT.}
  \label{fig:}
\end{figure*}
\begin{figure*}[h]
  \centering
  \includegraphics[width=0.8\linewidth]{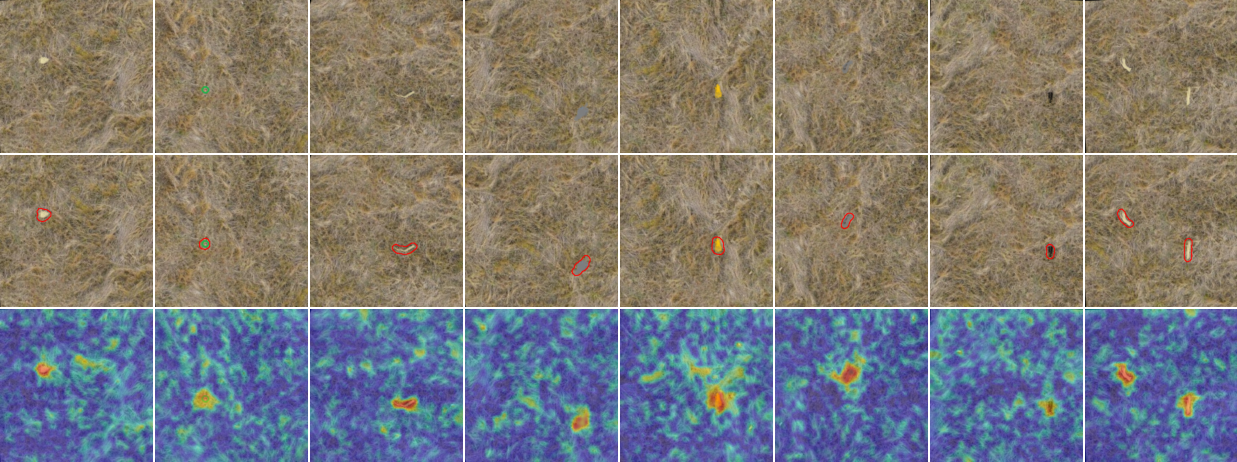}
  \caption{Visualization of input images (top), ground‐truth masks (middle), and model‐generated anomaly maps (bottom) for the \textit{Matted} category in the DTD dataset. All results are generated by PILOT.}
  \label{fig:}
\end{figure*}
\begin{figure*}[h]
  \centering
  \includegraphics[width=0.8\linewidth]{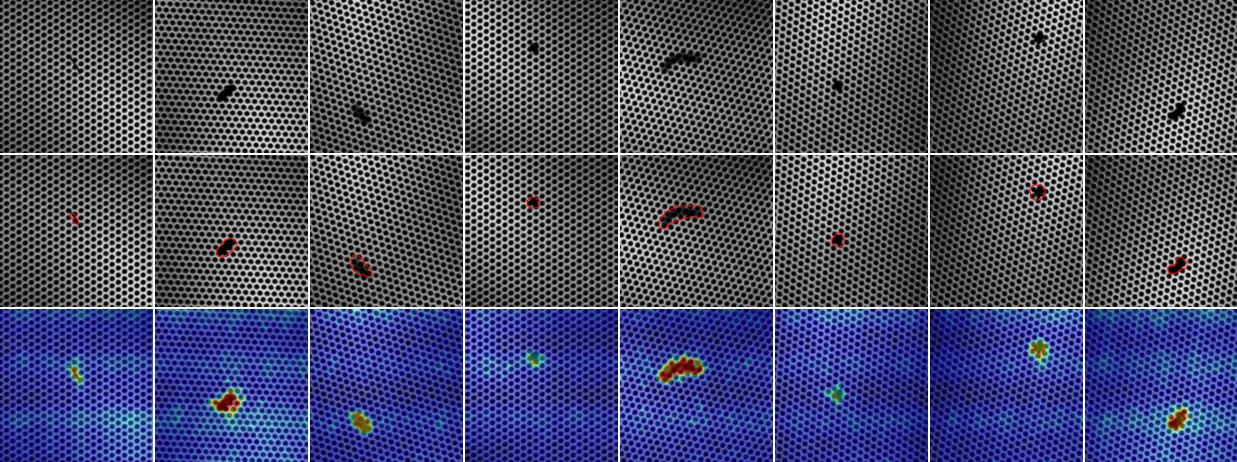}
  \caption{Visualization of input images (top), ground‐truth masks (middle), and model‐generated anomaly maps (bottom) for the \textit{Perforated} category in the DTD dataset. All results are generated by PILOT.}
  \label{fig:dtd_Perforated}
\end{figure*}

\end{document}